\documentclass[twoside,11pt]{article}

\usepackage{blindtext}

\usepackage[preprint]{jmlr2e}
\usepackage{hyperref}
\usepackage{url}
\usepackage{mathrsfs, graphicx, multirow}
\usepackage{caption}
\usepackage{subfigure}
\usepackage{amsmath,amsfonts,bm}
\usepackage{color}
\usepackage{amsmath}
\allowdisplaybreaks[4]

\usepackage{lastpage}
\jmlrheading{23}{2022}{1-\pageref{LastPage}}{1/21; Revised 5/22}{9/22}{21-0000}{Jiashuo Liu, Jiayun Wu, Bo Li and Peng Cui}

\ShortHeadings{Predictive Heterogeneity: Measures and Applications}{Jiashuo Liu, Jiayun Wu, Bo Li and Peng Cui}

\firstpageno{1}

\begin{document}

\title{Predictive Heterogeneity: Measures and Applications}

\author{\name Jiashuo Liu\thanks{Equal Contributions.} \email liujiashuo77@gmail.com \\
       \addr Department of Computer Science and Technology\\
       Tsinghua University
       \AND
       \name Jiayun Wu$^*$ \email jiayun.wu.work@gmail.com \\
       \addr Department of Computer Science and Technology\\
		Tsinghua University
		\AND 
	   \name Bo Li \email libo@sem.tsinghua.edu.cn\\
	   \addr School of Economics and Management\\
	   Tsinghua University
	   \AND 
	   \name Peng Cui\thanks{Corresponding Author.} \email cuip@tsinghua.edu.cn\\
	   \addr Department of Computer Science and Technology\\
	   Tsinghua University
}

\editor{My editor}

\maketitle

\begin{abstract}
As an intrinsic and fundamental property of big data, data heterogeneity exists in a variety of real-world applications, such as precision medicine, autonomous driving, financial applications, etc.
For machine learning algorithms, the ignorance of data heterogeneity will greatly hurt the generalization performance and the algorithmic fairness, since the prediction mechanisms among different sub-populations are likely to differ from each other.
In this work, we focus on the data heterogeneity that affects the prediction of machine learning models, and firstly propose the \emph{usable predictive heterogeneity}, which takes into account the model capacity and computational constraints.
We prove that it can be reliably estimated from finite data with probably approximately correct (PAC) bounds.
Additionally, we design a bi-level optimization algorithm to explore the usable predictive heterogeneity from data.
Empirically, the explored heterogeneity provides insights for sub-population divisions in income prediction, crop yield prediction and image classification tasks, and leveraging such heterogeneity benefits the out-of-distribution generalization performance.
\end{abstract}


\section{Introduction}


Big Data provides great opportunities for the growth and advancement of Artificial Intelligence (AI) systems. 
Nowadays, AI has emerged as a ubiquitous tool that permeates almost every aspect of the contemporary technological landscape, making it an indispensable asset in various fields and industries, such as scientific discoveries, policy-making, healthcare, drug discovery, and so on.
However, along with the widespread deployment of AI systems, the reliability, fairness, and stability of AI algorithms have been increasingly doubted.
For example, in sociological research \citep{2020Toward}, studies have shown that even for carefully designed randomized trials, there are huge selection biases, making scientific discoveries unreliable; in disease diagnosis, studies \citep{wynants2020prediction, roberts2021common} have found hundreds of existing AI algorithms fail to detect and prognosticate for COVID-19 using chest radiographs and CT scans; in social welfare, decision support AI systems for credit loan applications are found to exhibit biases against certain demographic groups \citep{hardt2016equality, verma2019weapons}; in various machine learning tasks, algorithms are faced with severely poor generalization performances under distributional shifts \citep{shen2021towards}, etc.
Another well-known example is Simpson's paradox, which brings false discoveries to the social research \citep{wagner1982simpson, hernan2011simpson}.

In order to mitigate the barriers that inhibit the deployment of AI systems in crucial, high-stakes applications, numerous researchers have taken recourse to the established research paradigm of model-centric AI, whereby they endeavor to develop innovative algorithms aimed at addressing these challenges.
However, in contemporary discourse about machine learning, it is increasingly evident that the challenges faced by algorithms extend beyond their intrinsic properties and extend to the nature of the data utilized in training these models. 
Specifically, the heterogeneity of data employed has emerged as a pivotal factor underlying these issues.
The concept of data heterogeneity encompasses the \emph{diversity} that exists within data, including \emph{variations in data sources, generation mechanisms, sub-populations}, and \emph{data structures}. 
Failure to account for such diversity in AI systems can lead to overemphasis on patterns found only in dominant sub-populations or groups, thereby resulting in false scientific discoveries, unreliable and inequitable decision-making, and poor generalization performance when confronted with new data. 
Given the high-stakes scenarios in which trustworthy AI is required, addressing the problem of data heterogeneity - an inherent property of big data - should receive increased attention. 
Moreover, in the current era of big models, where model development is approaching its limits, \emph{researchers have huge opportunities to explore the intricacies of big data}, thereby facilitating the development of AI in parallel with the advancement of AI models and algorithms.

Despite its widespread existence, due to its complexity, data heterogeneity has not converged to a uniform formulation so far, and has different meanings among different fields.
\cite{li1995definition} define the heterogeneity in \emph{ecology} based on the system property and complexity or variability.
\cite{rosenbaum2005heterogeneity} views the uncertainty of the potential outcome as unit heterogeneity in observational studies in \emph{economics}.
More recently, in machine learning, several works of \emph{causal learning} \citep{peters2016causal, arjovsky2019invariant, koyama2020out, liu2021heterogeneous,creager2021environment} and \emph{robust learning} \citep{sagawa2019distributionally, liu2022distributionally} leverage heterogeneous data from multiple environments to improve the out-of-distribution generalization ability.
\textcolor{black}{However, previous works have not provided a precise definition or sound quantification.
In this work, targeting at the prediction task in machine learning, from the perspective of \emph{prediction power}, we propose the predictive heterogeneity, a \emph{new type} of data heterogeneity.}


From a machine learning perspective, a major concern is the potential adverse effects of data heterogeneity on prediction accuracy. 
In this study, we propose predictive heterogeneity, which refers to the heterogeneity of data that impacts the performance of machine learning models. 
Our goal is to facilitate the development of machine learning systems by addressing this issue.
To this end, we introduce a precise definition of predictive heterogeneity that quantifies the maximal additional predictive information that can be obtained by dividing the entire data distribution into sub-populations. 
This measure takes into account the model capacity and computational constraints and can be accurately estimated from finite samples with probably approximately correct (PAC) bounds. 
We conduct a theoretical analysis of the properties of this measure and examine it under typical scenarios of data heterogeneity.
In addition, we propose the information maximization (IM) algorithm to empirically explore the predictive heterogeneity within data. 
Through our empirical investigations, we find that the explored heterogeneity is interpretable and provides valuable insights for sub-population divisions in various fields, such as agriculture, sociology, object recognition, and healthcare. 
Moreover, the identified sub-populations can be utilized to identify features related to Covid-19 mortality and enhance the out-of-distribution generalization performance of machine learning models. 
This has been confirmed through experiments with both simulated and real-world data.
In conclusion, our study contributes to the development of machine learning systems by providing a precise definition of predictive heterogeneity and a reliable measure for its estimation. 
Our findings demonstrate the potential of the IM algorithm for exploring predictive heterogeneity, assisting scientific discoveries and improving the generalization performance of machine learning models in real-world applications.

\section{Preliminaries on Mutual Information and Predictive $\mathcal V$-Information}
In this section, we briefly introduce the mutual information and predictive $\mathcal V$-information \citep{DBLP:conf/iclr/XuZSSE20} which are the preliminaries of our proposed predictive heterogeneity.\\\\
\textbf{Notations.} For a probability triple $(\mathbb S,  \mathcal F, \mathbb P)$, define random variables $X: \mathbb S\rightarrow \mathcal X$ and $Y: \mathbb S\rightarrow \mathcal Y$ where $\mathcal X$ is the covariate space and $\mathcal Y$ is the target space. Accordingly. $x \in \mathcal X$ denotes the covariates, and $y\in\mathcal{Y}$ denotes the target. Denote the set of random categorical variables as $\mathcal C = \{ C: \mathbb S \rightarrow \mathbb N| \text{supp}(C) \;\text{is finite} \}$. Additionally, $\mathcal{P}(\mathcal{X}), \mathcal{P}(\mathcal Y)$ denote the set of all probability measures over the Borel algebra on the spaces $\mathcal{X}, \mathcal{Y}$ respectively. 
$H(\cdot)$ denotes the Shannon entropy of a discrete random variable and the differential entropy of a continuous variable, and $H(\cdot|\cdot)$ denotes the conditional entropy of two random variables.

In information theory, the mutual information of two random variables $X$, $Y$ measures the dependence between the two variables, which quantifies the reduction of entropy for one variable when observing the other:
\begin{small}
\begin{equation}
	\mathbb{I}(X;Y) = H(Y) - H(Y|X).
\end{equation}	
\end{small}
It is known that the mutual information is associated with the predictability of $Y$ \citep{cover1991infomationtheory}. While the standard definition of mutual information unrealistically assumes the unbounded computational capacity of the predictor, rendering it hard to estimate especially in high dimensions.
To mitigate this problem, \cite{DBLP:conf/iclr/XuZSSE20} raise the predictive $\mathcal V$-information under realistic computational constraints, where the predictor is only allowed to use models in the predictive family $\mathcal V$ to predict the target variable $Y$.

\begin{definition}[Predictive Family \citep{DBLP:conf/iclr/XuZSSE20}]
	Let $\Omega=\{f:\mathcal{X}\cup\{\emptyset\}\rightarrow \mathcal{P}(\mathcal Y)\}$. We say that $\mathcal V \subseteq \Omega$ is a predictive family if it satisfies:
	\begin{small}
	\begin{equation}
	\label{equ:condition}
		\forall f\in\mathcal{V},\ \  \forall P\in \mathrm{range}(f),\ \  \exists f'\in\mathcal{V}, \quad\text{s.t. }\forall x\in\mathcal{X}, f'[x]=P, f'[\emptyset]=P.
	\end{equation}
	\end{small}
\end{definition}
A predictive family contains all predictive models that are allowed to use, which forms computational or statistical constraints.
The additional condition in Equation \ref{equ:condition} means that the predictor can always ignore the input covariates ($x$) if it chooses to (only use $\emptyset$).

\begin{definition}[Predictive $\mathcal V$-information \citep{DBLP:conf/iclr/XuZSSE20}]
\label{def:predictive_v_information}
	Let $X, Y$ be two random variables taking values in $\mathcal{X}\times\mathcal{Y}$ and $\mathcal V$ be a predictive family. The predictive $\mathcal V$-information from $X$ to $Y$ is defined as:
	\begin{small}
	\begin{equation}
		\mathbb{I}_{\mathcal V}(X\rightarrow Y) = H_{\mathcal V}(Y|\emptyset)-H_{\mathcal V}(Y|X),
	\end{equation}	
	\end{small}
	where $H_{\mathcal V}(Y|\emptyset)$, $H_{\mathcal V}(Y|X)$ are the predictive conditional $\mathcal V$-entropy defined as:
	\begin{small}
	\begin{align}
		H_{\mathcal V}(Y|X) &= \inf\limits_{f\in\mathcal{V}}\mathbb{E}_{x,y\sim X,Y}[-\log f[x](y)]. \\
		H_{\mathcal V}(Y|\emptyset) &= \inf\limits_{f\in\mathcal{V}}\mathbb{E}_{y\sim Y}[-\log f[\emptyset](y)].
	\end{align}	
	\end{small}
	Notably that $f\in\mathcal V$ is a mapping: $\mathcal{X}\cup\{\emptyset\}\rightarrow \mathcal{P}(\mathcal Y)$, so $f[x]\in\mathcal{P}(\mathcal{Y})$ is a probability measure on $\mathcal{Y}$, and $f[x](y)\in\mathbb{R}$ is the density evaluated on $y\in\mathcal Y$. $H_{\mathcal V}(Y|\emptyset)$ is also denoted as $H_{\mathcal V}(Y)$.
\end{definition}

Compared with the mutual information, the predictive $\mathcal V$-information restricts the computational power and is much easier to estimate in high-dimensional cases.
When the predictive family $\mathcal V$ contains all possible models, i.e. $\mathcal V = \Omega$, it is proved that $\mathbb{I}_{\mathcal V}(X\rightarrow Y)=\mathbb{I}(X;Y)$ \citep{DBLP:conf/iclr/XuZSSE20}.

\section{Predictive Heterogeneity}
In this paper, from the machine learning perspective, we quantify the data heterogeneity that affects decision making, named Predictive Heterogeneity, which is easy to integrate with machine learning algorithms and could help analyze big data and build more rational algorithms.

\subsection{Interaction Heterogeneity}
To formally define the predictive heterogeneity, we begin with the formulation of the interaction heterogeneity.
The \emph{interaction heterogeneity} is defined as:
\begin{definition}[Interaction Heterogeneity]
	Let $X$, $Y$ be random variables taking values in $\mathcal X \times \mathcal Y$. Denote the set of random categorical variables as $\mathcal C$, and take its subset $\mathscr E \subseteq \mathcal C$.  Then $\mathscr E$ is an environment set iff there exists $\mathcal E \in \mathscr E$ such that $X, Y \perp \!\!\! \perp \mathcal E$. $\mathcal E \in \mathscr E$ is called an environment variable. The interaction heterogeneity between $X$ and $Y$ w.r.t. the environment set $\mathscr E$ is defined as:
	\begin{small}
	\begin{equation}
	\label{equ:predictive-heterogeneity}
		\mathcal{H}^\mathscr E(X,Y) = \sup_{\mathcal{E} \in \mathscr E}\mathbb{I}(Y;X|\mathcal{E})-\mathbb{I}(Y;X).
	\end{equation} 
	\end{small}
\end{definition}
\vskip -0.1in

Each environment variable $\mathcal E$ represents a stochastic `partition' of $\mathcal X \times \mathcal Y$, and the condition for the environment set implies that there exists such a stochastic partition that the joint distribution of $X,Y$ could be preserved in each environment.
In information theory, $\mathbb{I}(Y;X|\mathcal{E})-\mathbb{I}(Y;X)$ is called the \emph{interaction information}, which measures the influence of the environment variable $\mathcal{E}$ on the amount of information shared between the target $Y$ and the covariate $X$.
And the \emph{interaction heterogeneity} defined in Equation \ref{equ:predictive-heterogeneity} quantifies the \emph{maximal} additional information that can be gained from involving or uncovering the environment variable $\mathcal{E}$.
Intuitively, large $\mathcal{H}^{\mathscr{E}}(X,Y)$ indicates that the predictive power from $X$ to $Y$ is enhanced by $\mathcal{E}$, which means that uncovering the latent sub-population associated with the environment partition $\mathcal{E}$ will benefit the $X\rightarrow Y$ prediction.

\subsection{Predictive Heterogeneity}
Based on the mutual information, the computation of the interaction heterogeneity is quite hard, since the standard mutual information is notoriously difficult to estimate especially in big data scenarios.
Also, even if the mutual information could be accurately estimated, the prediction model may not be able to make good use of it.

Inspired by \cite{DBLP:conf/iclr/XuZSSE20}, we raise the \emph{Predictive Heterogeneity}, which measures the interaction heterogeneity that can be captured under computational constraints and affects the prediction of models within the specified predictive family.
To begin with, we propose the \emph{Conditional Predictive $\mathcal V$-information}, which generalizes the predictive $\mathcal V$-information.

\begin{definition}[Conditional Predictive $\mathcal V$-information]
	Let $X, Y$ be two random variables taking values in $\mathcal{X}\times\mathcal{Y}$ and $\mathcal E$ be an environment variable. For a predictive family $\mathcal V$, the conditional predictive $\mathcal V$-information is defined as:
	\begin{equation}
		\mathbb{I}_{\mathcal{V}}(X\rightarrow Y|\mathcal{E}) = H_{\mathcal V}(Y|\emptyset,\mathcal E)-H_{\mathcal V}(Y|X,\mathcal E),
	\end{equation}	
	where $H_{\mathcal V}(Y|\emptyset,\mathcal E)$ and $H_{\mathcal V}(Y|X,\mathcal E)$ are defined as:
	\begin{align}
		H_{\mathcal V}(Y|X,\mathcal E) &= \mathbb E_{e \sim \mathcal E} \left[ \inf\limits_{f\in\mathcal{V}}\mathbb{E}_{x,y\sim X,Y|\mathcal E=e}[-\log f[x](y)]\right]. \\
		H_{\mathcal V}(Y|\emptyset,\mathcal E) &= \mathbb E_{e \sim \mathcal E} \left[ \inf\limits_{f\in\mathcal{V}}\mathbb{E}_{y\sim Y | \mathcal E=e}[-\log f[\emptyset](y)] \right].
	\end{align}	
\end{definition}

Intuitively, the conditional predictive $\mathcal V$-information measures the weighted average of predictive $\mathcal V$-information among environments.
And here we are ready to formalize the predictive heterogeneity measure.

\begin{definition}[Predictive Heterogeneity]
	Let $X$, $Y$ be random variables taking values in $\mathcal X \times \mathcal Y$ and $\mathscr E$ be an environment set. For a predictive family $\mathcal V$, the predictive heterogeneity for the prediction $X \rightarrow Y$ with respect to $\mathscr E$ is defined as:
	\begin{equation}
	\label{equ:usable-predictive-heterogeneity-1}
		\mathcal{H}^\mathscr E_{\mathcal V}(X \rightarrow Y) = \sup_{\mathcal{E} \in \mathscr E}\mathbb{I}_{\mathcal{V}}(X\rightarrow Y|\mathcal{E})-\mathbb{I}_{\mathcal{V}}(X\rightarrow Y),
	\end{equation} 
	where $\mathbb{I}_{\mathcal{V}}(X\rightarrow Y)$ is the predictive $\mathcal V$-information following from Definition \ref{def:predictive_v_information}.
\end{definition}

Leveraging the predictive $\mathcal V$-information, the predictive heterogeneity defined in Equation \ref{equ:usable-predictive-heterogeneity-1} characterizes the \emph{maximal additional information} that \emph{can be used} by the prediction model when involving the environment variable $\mathcal E$. 
It restricts the prediction models in $\mathcal{V}$ and the explored additional information could benefit the prediction performance of the model $f\in\mathcal V$, for which it is named predictive heterogeneity. 
Next, we present some basic properties of the interaction heterogeneity and predictive heterogeneity.

\begin{proposition}[\textcolor{black}{Basic Properties of Predictive Heterogeneity}]
\label{proposition1}
Let $X$, $Y$ be random variables taking values in $\mathcal X \times \mathcal Y$,  $\mathcal V$ be a function family, and $\mathscr E$, $\mathscr E_1$, $\mathscr E_2$   be environment sets.

    \quad 1. \emph{Monotonicity}: If $\mathscr E_1 \subseteq \mathscr E_2$, $\mathcal{H}^{\mathscr E_1}_{\mathcal V}(X \rightarrow Y) \leq \mathcal{H}^{\mathscr E_2}_{\mathcal V}(X \rightarrow Y)$.
    
    \quad 2. \emph{Nonnegativity}: $\mathcal{H}^{\mathscr E}_{\mathcal V}(X \rightarrow Y) \geq 0$.
    
    \quad 3. \emph{Boundedness}: For discrete $Y$, $\mathcal{H}^{\mathscr E}_{\mathcal V}(X \rightarrow Y) \leq H_\mathcal V(Y|X)$.
    
    \quad 4. \emph{Corner Case}: If the predictive family $\mathcal V$ is the largest possible predictive family that includes all possible models, i.e. $\mathcal V = \Omega$, we have $\mathcal{H}^\mathscr E(X,Y) = \mathcal{H}^{\mathscr E}_{\Omega}(X \rightarrow Y)$.
    
\end{proposition}

Proofs can be found at Appendix \ref{proof: prop1}.
For further theoretical properties of predictive heterogeneity, in Section \ref{sec:linear}, we derive its explicit forms under \emph{endogeneity}, a common reflection of data heterogeneity.
And we demonstrate in Section \ref{sec:bounds} that our proposed predictive heterogeneity can be empirically estimated with guarantees if the complexity of $\mathcal V$ is bounded (e.g., its Rademacher complexity).

\subsection{Theoretical Properties in Linear Cases}
\label{sec:linear}
In this section, we conduct a theoretical analysis of the predictive heterogeneity in multiple linear settings. Specifically, we consider two scenarios: (1) a homogeneous case with independent noises and (2) heterogeneous cases with endogeneity arising from selection bias and hidden variables. By examining these typical settings, we approximate the analytical forms of the proposed measure and draw insightful conclusions that can be generalized to more complex scenarios.

Firstly, under a homogeneous case with no data heterogeneity, Theorem \ref{theorem: homogeneous} proves that our measure is bounded by the scale of label noises (which is usually small) and reduces to 0 in linear case under mild assumptions.
It indicates that the predictive heterogeneity is insensitive to independent noises.
Notably that in the linear case we only deal with the environment variable satisfying $X\perp \epsilon|\mathcal E$, since in common prediction tasks, the independent noises are unknown and unrealistic to be exploited for the prediction.

\begin{theorem}[Homogeneous Case with Independent Noises]
\label{theorem: homogeneous}
	For a prediction task $X \rightarrow Y$ where $X$, $Y$ are random variables taking values in $\mathbb R^n \times \mathbb R$, 
	consider the data generation process as $Y = g(x) + \epsilon, \epsilon\sim\mathcal{N}(0,\sigma^2)$ where $g:\mathbb R^n \rightarrow  \mathbb R$ is a measurable function. 
	 1) For a function class $\mathcal G$ such that $g \in \mathcal G$, define the function family as $\mathcal V_{\mathcal G}=\{ f | f[x]=\mathcal{N}(\phi(x),\sigma_V^2), \phi \in \mathcal G, \sigma_V \in \mathbb R^+ \}$. With an environment set $\mathscr E$, we have $\mathcal{H}^\mathscr E_{\mathcal{V}_{\mathcal G}}(X \rightarrow Y)\leq \pi\sigma^2$.
	 2) Take $n=1$ and $g(x) =\beta x$,$\beta\in \mathbb R$. 
	 Without loss of generality, assume $\mathbb E[X] =0$ and $\mathbb E[X^2]$  exists. Given the function family $\mathcal{V}_\sigma=\{f | f[x]=\mathcal{N}(\theta x,\sigma^2), \theta \in \mathbb R, \sigma \text{ fixed }\}$ and the environment set $\mathscr E = \{ \mathcal  E | \mathcal E \in \mathcal C, |\text{supp}(\mathcal E)|=2, X \perp \epsilon | \mathcal E\}$. We have $\mathcal{H}^\mathscr E_{\mathcal{V}_\sigma}(X \rightarrow Y)=0$.
	 Proofs can be found at Appendix \ref{proof:homogeneous}.
\end{theorem}

Secondly, we examine the proposed measure under \emph{two typical cases of data heterogeneity} \citep{fan2014challenges}, named \emph{endogeneity by selection bias} \citep{heckman1979sample, winship1992models, cui2022stable} and \emph{endogeneity with hidden variables} \citep{fan2014challenges, arjovsky2019invariant}.

To begin with, in Theorem \ref{theorem: selection-bias}, we consider the prediction task $X \rightarrow Y$ with $X$, $Y$ taking values in $\mathbb R^2 \times \mathbb R$. Let $X = [S,V]^T$. The predictive family is specified as:
\begin{small} 
	\begin{equation}
	\label{equ:v_theorem23}
		\mathcal{V}=\{f|f[x]=\mathcal{N}(\theta_SS+\theta_VV, \sigma^2),\quad \theta_S,\theta_V\in \mathbb R, \sigma=1\}.
	\end{equation}
	\end{small}
And the data distribution $P(X,Y)$ is a mixture of latent sub-populations, which could be formulated by an environment variable $\mathcal E^* \in \mathcal C$ such that $P(X,Y) = \sum_{e\in \text{supp}(\mathcal E^*)} P(\mathcal E^* = e)P(X,Y|\mathcal E^* = e)$. For each $e \in \text{supp}(\mathcal E^*)$, $P(X,Y|\mathcal E^* = e)$ is the distribution of a homogeneous sub-population.
Note that the prediction task is to predict $Y$ with covariates $X$, and the sub-population structure is latent. That is, $P(\mathcal E^*|X,Y)$ is \emph{unknown} for models. In the following, we derive the analytical forms of our measure under the one typical case.

\begin{theorem}[Endogeneity with Selection Bias]
\label{theorem: selection-bias}
	For the prediction task $X=[S,V]^T\rightarrow Y$ with a latent environment variable $\mathcal E^*$, the data generation process with selection bias is defined as:
	\begin{small}
	\begin{align}
		Y = \beta S + f(S) + \epsilon_Y, \epsilon_Y \sim \mathcal{N}(0, \sigma_Y^2); \quad V = r(\mathcal E^*) f(S) + \sigma(\mathcal E^*)\cdot\epsilon_V, \epsilon_V \sim \mathcal{N}(0, 1),
	\end{align}
	\end{small}
	where  $f:\mathbb R\rightarrow \mathbb R$ and $r,\sigma:\text{supp}(\mathcal E^*) \rightarrow \mathbb R$ are measurable functions. $\beta \in \mathbb R$.
	Assume that $\mathbb{E}[S^2]$ is finite, $\mathbb{E}[f(S)S]=0$ and there exists $L>1$ such that $L\sigma^2(\mathcal E^*)< r^2(\mathcal E^*)\mathbb{E}[f^2]$. For the predictive family defined in equation \ref{equ:v_theorem23} and the environment set $\mathscr E = \mathcal C$, the predictive heterogeneity of the prediction task $[S,V]^T\rightarrow Y$ approximates to:
	\begin{small}
	\begin{equation}
	\label{equ:approximation1}
		\mathcal{H}^\mathcal C_{\mathcal{V}}(X\rightarrow Y) \approx \frac{\text{Var}(r_e)\mathbb{E}[f^2]+\mathbb{E}[\sigma^2(\mathcal E^*)]}{\mathbb{E}[r_e^2]\mathbb{E}[f^2]+\mathbb{E}[\sigma^2(\mathcal E^*)]}\mathbb{E}[f^2(S)], \text{error bounded by }\frac{1}{2}\max(\sigma_Y^2,R(r,\sigma,f)).
	\end{equation}	
	\end{small}
	And further we have
	\begin{equation}
		\begin{aligned}
			R(r(\mathcal E^*), \sigma(\mathcal E^*), f) &= \mathbb{E}[(\frac{1}{\frac{r^2\mathbb{E}[f^2]}{\sigma^2}+1})^2]\mathbb{E}[f^2]+ \mathbb{E}_{\mathcal{E}^*}[(\frac{1}{\frac{r}{\sigma}+\frac{\sigma}{r\mathbb{E}[f^2]}})^2]\\
			&< \mathbb{E}[f^2](\frac{1}{(L+1)^2}+\frac{1}{L+2+\frac{1}{L}}).
		\end{aligned}
	\end{equation}
	Proofs can be found at Appendix \ref{proof: linear}.
\end{theorem}

Intuitively, the data generation process in Theorem \ref{theorem: selection-bias} introduces the spurious correlation between the spurious feature $V$ and the target $Y$, which varies across different sub-populations (i.e. $r(\mathcal E^*)$ and $\sigma(\mathcal E^*)$ varies) and brings about data heterogeneity.
Here $\mathbb{E}[f(S)S]=0$ indicates a model misspecification since there is a nonlinear term $f(S)$ that could not be inferred by the linear predictive family with the stable feature $S$. 
The constant $L$ characterizes the strength of the spurious correlation between $V$ and $Y$.
Larger $L$ means $V$ could provide more information for prediction.

From the approximation in Equation \ref{equ:approximation1}, we can see that our proposed predictive heterogeneity is dominated by two terms: (1) $\text{Var}[r(\mathcal E^*)]/\mathbb{E}[r^2(\mathcal E^*)]$ characterizes the variance of $r(\mathcal E^*)$ among sub-populations; (2) $\mathbb{E}[f^2(S)]$ reflects the strength of model misspecifications.
These two components account for two sources of the data heterogeneity under selection bias, which validates the rationality of our proposed measure. 
Based on the theorem, it can be inferred that the degree of predictive heterogeneity increases with greater variability of $r(\mathcal E^*)$ among sub-populations and stronger model misspecifications. 
In other words, when the sub-populations differ significantly from each other and the model is not accurately specified, the predictive heterogeneity is likely to be larger.

 Additionally, in Theorem \ref{theorem: omitted variable}, we analyze our measure under endogeneity with hidden variables.
 In Theorem \ref{theorem: omitted variable}, an anti-causal covariate $V$ is generated via the causal diagram  like $Y\rightarrow V \leftarrow \mathcal E^*$ with a hidden environment variable $\mathcal E^*$.
 However, since $\mathcal E^*$ is omitted from the prediction models, the relationship between $V$ and $Y$ is biased, which inhibits the prediction power.

 \begin{theorem}[Endogeneity with Hidden Variables]
 \label{theorem: omitted variable}
 	For the prediction task $[S,V]^T\rightarrow Y$ with a latent environment variable $\mathcal E^*$, the data generation process with hidden variables is defined as:
 	\begin{small}
 	\begin{align}
 		Y = \beta S + f(S) + \epsilon_Y, \epsilon_Y\sim\mathcal{N}(0, \sigma_Y^2); \quad V = r(\mathcal E^*) (f(S)+\epsilon_Y) + \sigma(\mathcal E^*)\epsilon_V, \epsilon_V \sim\mathcal{N}(0, 1),
 	\end{align}
 	\end{small}
 	where  $f:\mathbb R\rightarrow \mathbb R$ and $r,\sigma:\text{supp}(\mathcal E^*) \rightarrow \mathbb R$ are measurable functions. $\beta \in \mathbb R$.
 	Assume that $\mathbb{E}[f(S)S]=0$ and there exists $L>1$ such that $L\sigma^2(\mathcal E^*) < r^2(\mathcal E^*)(\mathbb{E}[f^2]+\sigma_Y^2)$. 
 	For the predictive family defined in equation \ref{equ:v_theorem23} and the environment set $\mathscr E = \mathcal C$, the predictive heterogeneity of the prediction task $[S,V]^T\rightarrow Y$ approximates to:
 	\begin{small}
	\begin{equation}
	\begin{aligned}
	\label{equ:approximation2}
		&\mathcal{H}^\mathcal C_{\mathcal{V}}(X\rightarrow Y) \approx \frac{\text{Var}(r_e)(\mathbb{E}[f^2]+\sigma_Y^2)+\mathbb{E}[\sigma^2(\mathcal E^*)]}{\mathbb{E}[r_e^2](\mathbb{E}[f^2]+\sigma_Y^2)+\mathbb{E}[\sigma^2(\mathcal E^*)]}(\mathbb{E}[f^2(S)]+\sigma_Y^2),\\
		&\text{error bounded by }\frac{1}{2}\max(\sigma_Y^2,R(r,\sigma,f)).
	\end{aligned}
	\end{equation}	
	\end{small}
	And further we have:
	\begin{small}
	\begin{equation}
		\begin{aligned}
			R(r(\mathcal E^*), \sigma(\mathcal E^*), f) &= \mathbb{E}[(\frac{1}{\frac{r^2(\mathbb{E}[f^2]+\sigma_Y^2)}{\sigma^2}+1})^2](\mathbb{E}[f^2]+\sigma_Y^2)+ \mathbb{E}_{\mathcal{E}^*}[(\frac{1}{\frac{r}{\sigma}+\frac{\sigma}{r(\mathbb{E}[f^2]+\sigma_Y^2)}})^2]\\
			&<(\mathbb{E}[f^2]+\sigma_Y^2)(\frac{1}{(L+1)^2}+\frac{1}{L+2+\frac{1}{L}}).
		\end{aligned}
	\end{equation}	
	\end{small}
	Proofs can be found at Appendix \ref{proof: linear}.
 \end{theorem}

Intuitively, the data generation process in Theorem \ref{theorem: omitted variable} introduces the \emph{biased} anti-causal relationship between the spurious feature $V$ and the target $Y$, which varies across different sub-populations (i.e. $r(\mathcal E^*)$ and $\sigma(\mathcal E^*)$ varies) and brings about data heterogeneity.
Here, similar as Theorem \ref{theorem: selection-bias}, $\mathbb{E}[f(S)S]=0$ indicates model misspecification and the constant $L$ characterizes the strength of the biased anti-causal relationship between $V$ and $Y$, where larger $L$ means more information that $V$ could provide for predicting $Y$ when $\mathcal E^*$ is missing.
From the approximation in Equation \ref{equ:approximation2}, we can see that our proposed predictive heterogeneity is dominated by two terms: (1) $\text{Var}[r(\mathcal E^*)]/\mathbb{E}[r^2(\mathcal E^*)]$ characterizes the variance of $r(\mathcal E^*)$ among sub-populations; (2) $\mathbb{E}[f^2(S)]+\sigma_Y^2$ reflects the maximal additional information that could be provided by $V$.

In the broader context, Theorem 1, 2, and 3 suggest that our proposed predictive heterogeneity measure is equipped with remarkable properties, namely its insensitivity to homogeneous cases and its ability to account for the latent heterogeneity arising from typical sources of data heterogeneity. 
These findings highlight the efectiveness of our measure in accurately characterizing predictive heterogeneity in various machine learning tasks.

\subsection{PAC Guarantees for Predictive Heterogeneity Estimation}
Defined under explicit computation constraints, our Predictive Heterogeneity could be empirically estimated with guarantees if the complexity of the model family $\mathcal V$ is bounded.
In this work, we provide finite sample generalization bounds with the Rademacher complexity. First, we describe the definition of the empirical predictive heterogeneity, the explicit formula for which could be found in Definition \ref{def:empirical_predictive_heterogeneity}.

The dataset $\mathcal D = \{(x_i,y_i)\}_{i=1}^{|\mathcal D|}$ is independently and identically drawn from the population $X,Y$. 
Given a function family $\mathcal V $ and an environment set $\mathscr E_K$ such that for $\mathcal E\in \mathscr E_K$, $\text{supp}(\mathcal E) = \{ (e_k)_{k=1}^K \}$. , let $\mathcal Q$ be the set of all probability distributions of $X$,$Y$,$\mathcal E$ where $\mathcal E \in \mathscr E_K$.  The empirical predictive heterogeneity $\hat{\mathcal H}_{\mathcal V}^{\mathscr E_K}(X \rightarrow Y; \mathcal D)$ is given by:
\begin{small}
\begin{align}
    \hat{\mathcal H}_{\mathcal V}^{\mathscr E_K}(X \rightarrow Y; \mathcal D) 
    &= \sup_{\mathcal E \in \mathscr E_K }\hat{\mathbb{I}}_{\mathcal{V}}(X\rightarrow Y|\mathcal{E};\mathcal D)-\hat {\mathbb{I}}_{\mathcal{V}}(X\rightarrow Y;\mathcal D) \\
    &= \sup_{\hat Q \in \mathcal Q } {\sum_{k=1}^K \left[\hat Q(\mathcal E=e_k)\hat H_{\mathcal V}(Y|\mathcal E=e_k;\mathcal D) - \hat Q(\mathcal E=e_k)\hat H_{\mathcal V}(Y|X, \mathcal E=e_k;\mathcal D)\right]} \\
    & \quad\quad - [\hat H_{\mathcal V}(Y;\mathcal D) - \hat H_{\mathcal V}(Y|X;\mathcal D)].
\end{align}
\end{small}
Specifically, 
\begin{align}
    & \quad\; \hat Q(\mathcal E=e_k)\hat H_{\mathcal V}(Y|X, \mathcal E=e_k;\mathcal D) \\
    &= \inf_{f \in \mathcal V} \hat Q(\mathcal E=e_k) \sum_{x_i,y_i \in \mathcal D} -\log  f[x_i](y_i) \frac{\hat Q(x_i,y_i|\mathcal E=e_k)}{\sum_{x_j,y_j \in \mathcal D}\hat Q(x_j,y_j|\mathcal E=e_k) }  \\
    &= \inf_{f \in \mathcal V} \hat Q(\mathcal E=e_k) \sum_{x_i,y_i \in \mathcal D} -\log f[x_i](y_i) \frac{\hat Q(\mathcal E=e_k|x_i,y_i)
    \hat Q(x_i,y_i)}{\sum_{x_j,y_j \in \mathcal D}\hat Q(\mathcal E=e_k|x_j,y_j)\hat Q(x_j,y_j) } \\
    &= \inf_{f \in \mathcal V} \hat Q(\mathcal E=e_k) \sum_{x_i,y_i \in \mathcal D} -\log f[x_i](y_i) \frac{\hat Q(\mathcal E=e_k|x_i,y_i)
    \hat Q(x_i,y_i)}{\hat Q(\mathcal E=e_k) } \\
    &= \inf_{f \in \mathcal V} \sum_{x_i,y_i \in \mathcal D} -\log f[x_i](y_i) \hat Q(\mathcal E=e_k|x_i,y_i) \hat Q(x_i,y_i) \\
    &= \inf_{f \in \mathcal V} \frac{1}{|\mathcal D|}  \sum_{x_i,y_i \in \mathcal D} -\log f[x_i](y_i) \hat Q(\mathcal E=e_k|x_i,y_i).
\end{align}

The explicit formula for $\hat Q(\mathcal E=e_k)\hat H_{\mathcal V}(Y|\mathcal E=e_k;\mathcal D)$, $\hat H_{\mathcal V}(Y|X;\mathcal D)$ and $\hat H_{\mathcal V}(Y;\mathcal D)$ could be similarly derived. Here we are ready to formally define the empirical predictive heterogeneity.

\begin{definition} [Empirical Predictive Heterogeneity]
\label{def:empirical_predictive_heterogeneity}
	For the prediction task $X\rightarrow Y$ with $X$, $Y$ taking values in $\mathcal X \times \mathcal Y$, a dataset $\mathcal D$ is independently and identically drawn from the population such that $\mathcal D=\{(x_i,y_i)_{i=1}^N \sim X,Y \}$. 
	Given the predictive family $\mathcal V$ and the environment set $\mathscr E_K =\{ \mathcal E|\mathcal E \in \mathcal C, |\text{supp}(\mathcal E)|=K \}$ where $K \in \mathbb N$.
	Without loss of generality, we specify that  $\text{supp}(\mathcal E) = \{ (e_k)_{k=1}^K \}$ where $e_k$ denotes a single environment.
	Let $\mathcal Q$ be the set of all probability distributions of $X$,$Y$,$\mathcal E$ where $\mathcal E \in \mathscr E_K$. The empirical predictive heterogeneity $\hat{\mathcal{H}}^{\mathscr{E}_K}_{\mathcal V}(X\rightarrow Y; \mathcal D)$ with respect to $\mathcal D$  is defined as:
	\begin{equation}
	\label{equ:appendix-formal-empirical}
	    \begin{aligned}
    \hat{\mathcal H}_{\mathcal V}^{\mathscr E_K}(X \rightarrow Y; \mathcal D) 
    &= \sup_{\hat Q \in \mathcal Q } {\sum_{k=1}^K \left[\hat Q(\mathcal E=e_k)\hat H_{\mathcal V}(Y|\mathcal E=e_k;\mathcal D) - \hat Q(\mathcal E=e_k)\hat H_{\mathcal V}(Y|X, \mathcal E=e_k;\mathcal D)\right]} \\
    & \quad\quad - [\hat H_{\mathcal V}(Y;\mathcal D) - \hat H_{\mathcal V}(Y|X;\mathcal D)],
\end{aligned}
	\end{equation}
where
\begin{align}
    \hat Q(\mathcal E=e_k)\hat H_{\mathcal V}(Y|X, \mathcal E=e_k;\mathcal D) &=
    \inf_{f \in \mathcal V} \frac{1}{|\mathcal D|}  \sum_{x_i,y_i \in \mathcal D} -\log f[x_i](y_i) \hat Q(\mathcal E=e_k|x_i,y_i). \\
    \hat Q(\mathcal E=e_k)\hat H_{\mathcal V}(Y|\mathcal E=e_k;\mathcal D) &= \inf_{f \in \mathcal V} \frac{1}{|\mathcal D|}  \sum_{x_i,y_i \in \mathcal D} -\log f[\emptyset](y_i) \hat Q(\mathcal E=e_k|x_i,y_i). \\
    \hat H_{\mathcal V}(Y|X;\mathcal D) &=\inf_{f \in \mathcal V} \frac{1}{|\mathcal D|}  \sum_{x_i,y_i \in \mathcal D} -\log  f[x_i](y_i). \\
    \hat H_{\mathcal V}(Y;\mathcal D) &=\inf_{f \in \mathcal V} \frac{1}{|\mathcal D|}  \sum_{x_i,y_i \in \mathcal D} -\log  f[\emptyset](y_i).
\end{align}
\end{definition}


Then we give the PAC bound over the empirical usable predictive heterogeneity in Theorem \ref{theorem:pac}.
\label{sec:bounds}
\begin{theorem}[PAC Bound]
\label{theorem:pac}
	Consider the prediction task $X \rightarrow Y$ where $X$, $Y$ are random variables taking values in $\mathcal X \times \mathcal Y$. Assume that the predictive family $\mathcal V$ satisfies $\forall x\in\mathcal{X}$, $\forall y \in \mathcal Y$,$\forall f \in \mathcal V$, $\log f[x](y) \in [-B,B]$ where $B > 0$.  
	For given $K \in \mathbb N$, the environment set is defined as $\mathscr E_K = \{ \mathcal E| \mathcal E \in \mathcal C, |\text{supp}(\mathcal E)| = K \}$ where $K \in \mathbb N$. 
	Without loss of generality, we specify that  $\text{supp}(\mathcal E) = \{ (e_k)_{k=1}^K \}$ where $e_k$ denotes a single environment.
	Let $\mathcal Q$ be the set of all probability distributions of $X$,$Y$,$\mathcal E$ where $\mathcal E \in \mathscr E_K$.  Take an $e \in \text{supp}(\mathcal E)$ and define a function class $\mathcal G_{\mathcal V} = \{g|g(x,y) = \log f[x](y)Q(\mathcal E=e|x,y), f\in \mathcal V, Q \in \mathcal Q  \}$. Denote the Rademacher complexity of $\mathcal{G}$ with $N$ samples by $\mathscr{R}_{N}(\mathcal{G})$.
Then for any $\delta \in \left(0,1/(2K+2)\right)$, with a probability over $1 - 2(K+1)\delta$, for dataset $\mathcal{D}$ independently and identically drawn from $X$, $Y$, we have:
\begin{small}
\begin{align}
    |\mathcal{H}^{\mathscr E_K}_{\mathcal V}(X\rightarrow Y) - \hat{\mathcal H}^{\mathscr E_K}_{\mathcal V}(X\rightarrow Y;\mathcal D)| \leq 4(K+1)\mathscr R_{|\mathcal D|}(\mathcal G_{\mathcal V}) + 2(K+1)B\sqrt{2\log{\frac{1}{\delta}}/|\mathcal D|},
\end{align}
\end{small}
where $\mathscr R_{|\mathcal D|}(\mathcal G_{\mathcal V}) = \mathcal O(|\mathcal{D}|^{-\frac{1}{2}})$ \citep{bartlett2002rademacher}.
Proofs can be found at Appendix \ref{proof: pac}.
\end{theorem}

\section{Algorithm}

To empirically estimate the predictive heterogeneity in Definition \ref{def:empirical_predictive_heterogeneity}, we derive the Information Maximization (IM) algorithm from the formal definition in Equation \ref{equ:appendix-formal-empirical} to infer the distribution of $\mathcal{E}$ that maximizes the empirical predictive heterogeneity $\hat{\mathcal{H}}^{\mathscr{E}_K}_{\mathcal V}(X\rightarrow Y; \mathcal D)$.

\textbf{Objective Function.}\quad Given dataset $\mathcal D=\{X_N,Y_N\} = \{(x_i,y_i)\}_{i=1}^N$, denote $\text{supp}(\mathcal{E})=\{e_1, \dots, e_K\}$, we parameterize the distribution of $\mathcal{E}|(X_N,Y_N)$ with weight matrix $W\in\mathcal{W}_K$, where $K$ is the pre-defined number of environments and $\mathcal{W}_K=\{W: W\in\mathbb{R}_+^{N\times K}\text{ and }W\textbf{1}_K=\textbf{1}_N\}$ is the allowed weight space.
Each element $w_{ij}$ in $W$ represents $P(\mathcal{E}=e_j|x_i,y_i)$ (the probability of the $i$-th data point belonging to the $j$-th sub-population).
For a predictive family $\mathcal V$, the solution to the supremum problem in the Definition \ref{def:empirical_predictive_heterogeneity} is equivalent to the following objective function:
\begin{small}
\begin{equation}
\label{equ:bi-level}
\begin{aligned}
	\min\limits_{W\in\mathcal{W}_K} &\mathcal{R}_{\mathcal V}(W,\theta_1^*(W),\dots,\theta^*_K(W))=\left\{\frac{1}{N}\sum_{i=1}^N\sum_{j=1}^K w_{ij}\ell_{\mathcal V}(f_{\theta_j^*}(x_i),y_i) + U_{\mathcal V}(W,Y_N)
	\right\},\\
	&\text{s.t.}\quad  \theta^*_j(W) \in \arg\min_{\theta} \left\{\mathcal{L}_{\mathcal V}^j(W,\theta)=\sum_{i=1}^Nw_{ij}\ell_{\mathcal V}(f_{\theta}(x_i),y_i)\right\},\quad \text{for }j=1,\dots,K,
\end{aligned}		
\end{equation}
\end{small}
where $f_{\theta}:\mathcal{X}\rightarrow \mathcal{Y}$ denotes a predicting function parameterized by $\theta$, $\ell_{\mathcal{V}}(\cdot,\cdot):\mathcal{Y}\times\mathcal{Y}\rightarrow \mathbb{R}$ represents a loss function and $U_{\mathcal V}(W,Y_N)$ is a regularizer.
Specifically, $f_\theta$, $\ell_{\mathcal V}$ and $U_{\mathcal V}$ are determined by the predictive family $\mathcal{V}$. 
Here we provide implementations for two typical and general machine learning tasks, regression and classification.

\subsection{Regression}
For the \emph{regression task}, the predictive family is typically modeled as:
\begin{small}
\begin{equation}
	\mathcal{V}_1 = \{g: g[x]=\mathcal{N}(f_{\theta}(x), \sigma^2), f\text{ is the predicting function and }\theta\text{ is learnable, }\sigma \text{ is a constant}\}.
\end{equation}	
\end{small}
The corresponding loss function is $\ell_{\mathcal{V}_1}(f_\theta(X),Y)=(f_\theta(X)-Y)^2$, and $U_{\mathcal{V}_1}(W,Y_N)$ becomes 
\begin{small}
\begin{equation}
\label{equ:regularizer-regression}
	U_{\mathcal{V}_1}(W,Y_N) = \text{Var}_{j\in [K]}(\overline{Y_N^j})= \sum_{j=1}^K\left(\sum_{i=1}^N w_{ij}y_i\right)^2\frac{1}{N\sum_{i=1}^Nw_{ij}}-\left(\frac{1}{N}\sum_{i=1}^Ny_i\right)^2
\end{equation}	
\end{small}
where $\overline{Y^j_N}$ denotes the mean value of the label $Y$ given $\mathcal{E}=e_j$ and $U(W,Y_N)$ calculates the variance of $\overline{Y^j_N}$ among sub-populations $e_1\sim e_K$.

\subsection{Classification}
For the \emph{classification task}, the predictive family is typically modeled as:
\begin{small}
\begin{equation}
	\mathcal{V}_2 = \{g: g[x]=f_\theta(x)\in\Delta_c, f\text{ is the classification model and }\theta\text{ is learnable}\},
\end{equation}	
\end{small}
where $c$ is the class number and $\Delta_c$ denotes the $c$-dimensional simplex.
Here each model in the predictive family $\mathcal V_2$ outputs a discrete distribution in the form of a $c$-dimensional simplex.
In this case, the corresponding loss function $\ell_{\mathcal V_2}(\cdot,\cdot)$ is the cross entropy loss and the regularizer becomes $U_{\mathcal V_2}(W,Y_N) = -\sum_{j=1}^K \frac{1}{N}(\sum_{i=1}^Nw_{ij}) H(Y_N^j)$, where $H(Y_N^j)$ is the entropy of $Y$ given $\mathcal{E}=e_j$.


\subsection{Optimization.}
The bi-level optimization in Equation \ref{equ:bi-level} can be solved by performing projected gradient descent w.r.t. $W$.
The gradient of $W$ can be calculated by: (we omit the subscript $\mathcal{V}$ here)
\begin{small}
\begin{align}
	\nabla_W\mathcal R &=\nabla_W U + \left[\ell(f_{\theta_j}(x_i),y_i)\right]_{i,j}^{N\times K} + \sum_{j=1}^K \boxed{\nabla_{\theta_j}\mathcal{R}|_{\theta^*_j}\nabla_W\theta_j^*},\\
	\label{equ:tstep}
	\text{where }\boxed{\left.\nabla_{\theta_j}\mathcal{R}\right\vert_{\theta^*_j}\nabla_W\theta_j^*} & \approx \left.\nabla_{\theta_j}\mathcal{R}\right\vert_{\theta_j^t}\sum_{h\leq t}\left [ \prod_{k<h}(I - \left.\frac{\partial^2\mathcal{L}^j}{\partial\theta_j \partial \theta_j^{\mathrm{T}}}\right\vert_{\theta_j^{t-k-1}})\right ]\left.\frac{\partial^2\mathcal{L}^j}{\partial\theta_j\partial W^{\mathrm{T}}}\right\vert_{\theta_j^{t-h-1}} \\
	\label{equ:1step}
	&\approx  \left.\nabla_{\theta_j}\mathcal{R}\right\vert_{\theta_j^t}\left.\frac{\partial^2\mathcal{L}^j}{\partial \theta_j\partial W^{\mathrm{T}}}\right \vert_{\theta_j^{t-1}}\quad\quad\text{  , for }j=1,\dots,K.
\end{align}
\end{small}
where $\ell(f_{\theta_j}(x_i),y_i)]_{i,j}^{N\times K}$ is an $N\times K$ matrix in which the $(i,j)$-th element is $\ell(f_{\theta_j}(x_i),y_i)$.
Here Equation \ref{equ:tstep} approximates $\theta_j^*$ by $\theta_j^t$ from $t$ steps of inner loop gradient descent and Equation \ref{equ:1step} takes $t=1$ and performs \emph{1-step truncated backpropagation} \citep{shaban2019truncated,zhou2022model}.
Our information maximization algorithm updates $W$ by projected gradient descent as:
\begin{small}
\begin{equation}
	W \leftarrow \text{Proj}_{\mathcal{W}_K}\left(W-\eta\nabla_W\mathcal R\right),\quad \eta\text{ is the learning rate of }W.
\end{equation}	
\end{small}

Then we prove that minimizing Equation \ref{equ:bi-level} exactly finds the supremum w.r.t. $\mathcal{E}$ in the Definition \ref{def:empirical_predictive_heterogeneity} (formal) of the empirical predictive heterogeneity.

\begin{theorem}[Justification of the IM Algorithm]
\label{theorem:IM}
	For the regression task with predictive family $\mathcal{V}_1$ and classification task with $\mathcal{V}_2$, the optimization of Equation \ref{equ:bi-level} is equivalent to the supremum problem of the empirical predictive heterogeneity $\hat{\mathcal{H}}^{\mathscr{E}_K}_{\mathcal V_1}(X\rightarrow Y; \mathcal D)$, $\hat{\mathcal{H}}^{\mathscr{E}_K}_{\mathcal V_2}(X\rightarrow Y; \mathcal D)$ respectively in Equation \ref{equ:appendix-formal-empirical}  with the pre-defined environment number $K$ (i.e. $|\text{supp}(\mathcal E)|=K$).
	Proofs can be found at Appendix \ref{proof: IM}.
\end{theorem}

\begin{remark}[Difference from Expectation Maximization]
	The expectation maximization (EM) algorithm is to infer latent variables of a statistic model to achieve the \textbf{maximum likelihood}.
	Our proposed information maximization (IM) algorithm is to infer the latent variables $W$ which brings the \textbf{maximal predictive heterogeneity} associated with the maximal information.
	Due to the regularizer $U_{\mathcal V}$ in our objective function, the EM algorithm cannot efficiently solve our problem, and therefore we adopt bi-level optimization techniques. 
\end{remark}


\subsection{Approximation Accuracy}
\begin{figure}[t]
    \centering
    \includegraphics[width=\textwidth]{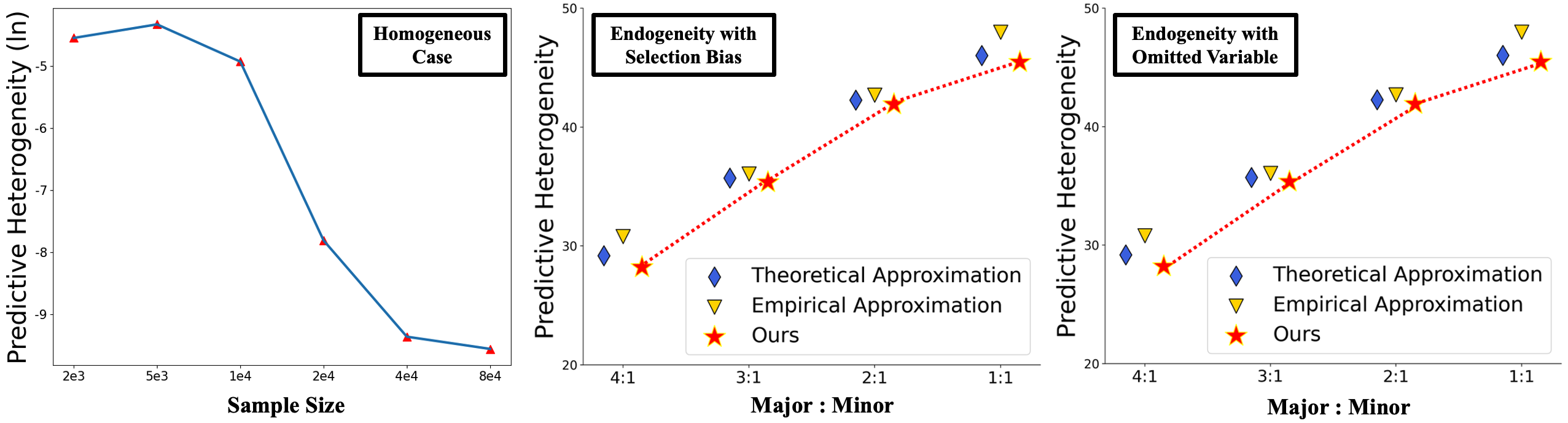}
    \caption{Numerical results of the toy examples in Section \ref{sec:linear}. The left sub-figure plots the estimated predictive heterogeneity under the setting of Theorem \ref{theorem: homogeneous}, the middle sub-figure plots the theoretical approximation, empirical approximation and our results under the setting of Theorem \ref{theorem: selection-bias}, and the right one is under the setting of Theorem \ref{theorem: omitted variable}.}
    \label{fig:appendix-toy}
\end{figure}

Here we provide some additional numerical results of our linear examples in Section \ref{sec:linear}.
In the left sub-figure of Figure \ref{fig:appendix-toy}, we plot the estimated predictive heterogeneity under the setting of Theorem \ref{theorem: homogeneous} where the analytical solution is equal to 0.
From the results, we can see that with the growing of sample size, the estimated value of our IM algorithm is approaching to 0 (note that the $y$-axis is $\ln(\text{estimated value})$).
In the middle sub-figure, for the setting in Theorem \ref{theorem: selection-bias}, we plot the theoretical approximation, empirical approximation (finite sample case) and the estimated value of the predictive heterogeneity under different ratios between the majority and the minority (which controls the $\text{Var}[r(\mathcal{E}^*)]$ in Equation \ref{equ:approximation1}).
And the right sub-figure plots the same values under the setting in Theorem \ref{theorem: omitted variable}.
From these two figures, we can see that (1) the empirical approximation under finite samples lies closely to the theoretical approximation, which is supported by our generalization bounds in Theorem \ref{theorem:pac}; (2) the estimated value of our IM algorithm is closely to the theoretical approximation,, which demonstrates the accuracy of our approximation algorithm in Equation \ref{equ:tstep} and \ref{equ:1step}. 
Also, as the ratio changes from $4:1$ to $1:1$, the data heterogeneity is increasing, and our predictive heterogeneity is also increasing, which is controlled by the term $\text{Var}(r(\mathcal{E}^*))$ in Equation \ref{equ:approximation1} and \ref{equ:approximation2}.

\section{Experiments}
\label{section:exp}

\begin{figure}[b]
  \centering
  \includegraphics[width=\textwidth,height=4.5cm]{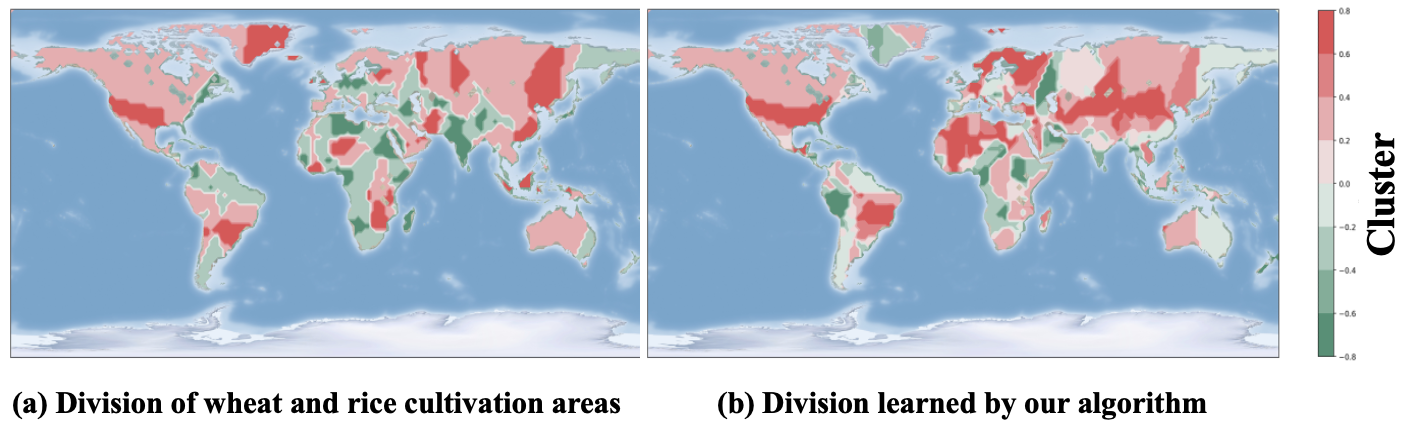}
  \vskip -0.1in
  \caption{Results on the crop yield data. We color each region according to its main crop type, and the shade represents the proportion of the main crop type after smoothing via $k$-means ($k=3$).}
  \label{fig:climate}
  \vskip -0.2in
\end{figure}

\begin{figure}[t]
\centering
\begin{minipage}{.48\textwidth}
  \centering
  \includegraphics[width=\linewidth]{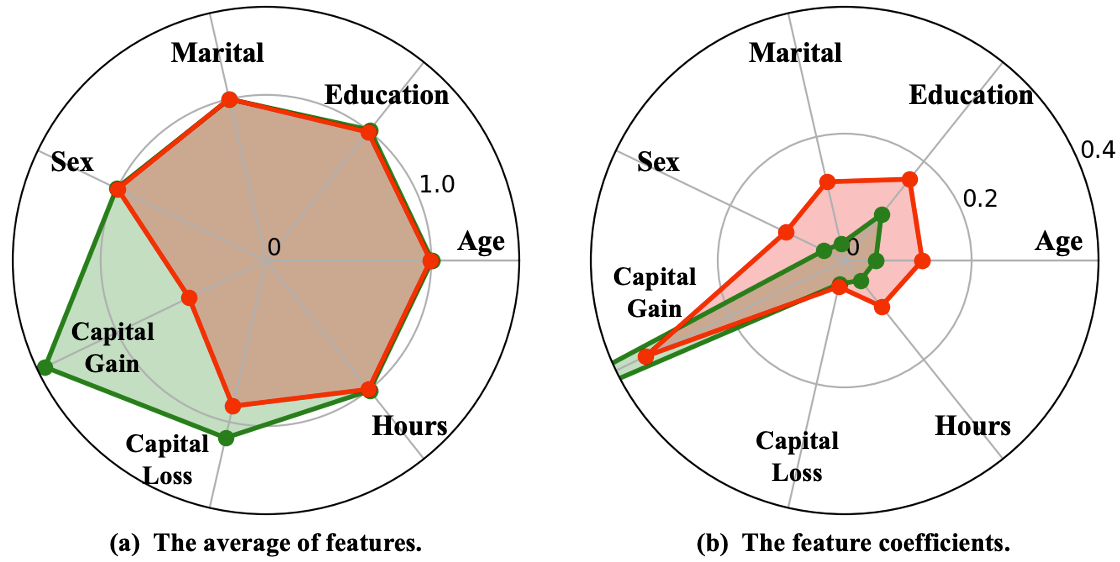}
  \caption{Results on the Adults data. Here we show the average of features and the feature coefficients of the two learned sub-populations.}
  \label{fig:test1}
\end{minipage}%
\hfill
\begin{minipage}{.48\textwidth}
  \centering
  \includegraphics[width=\linewidth]{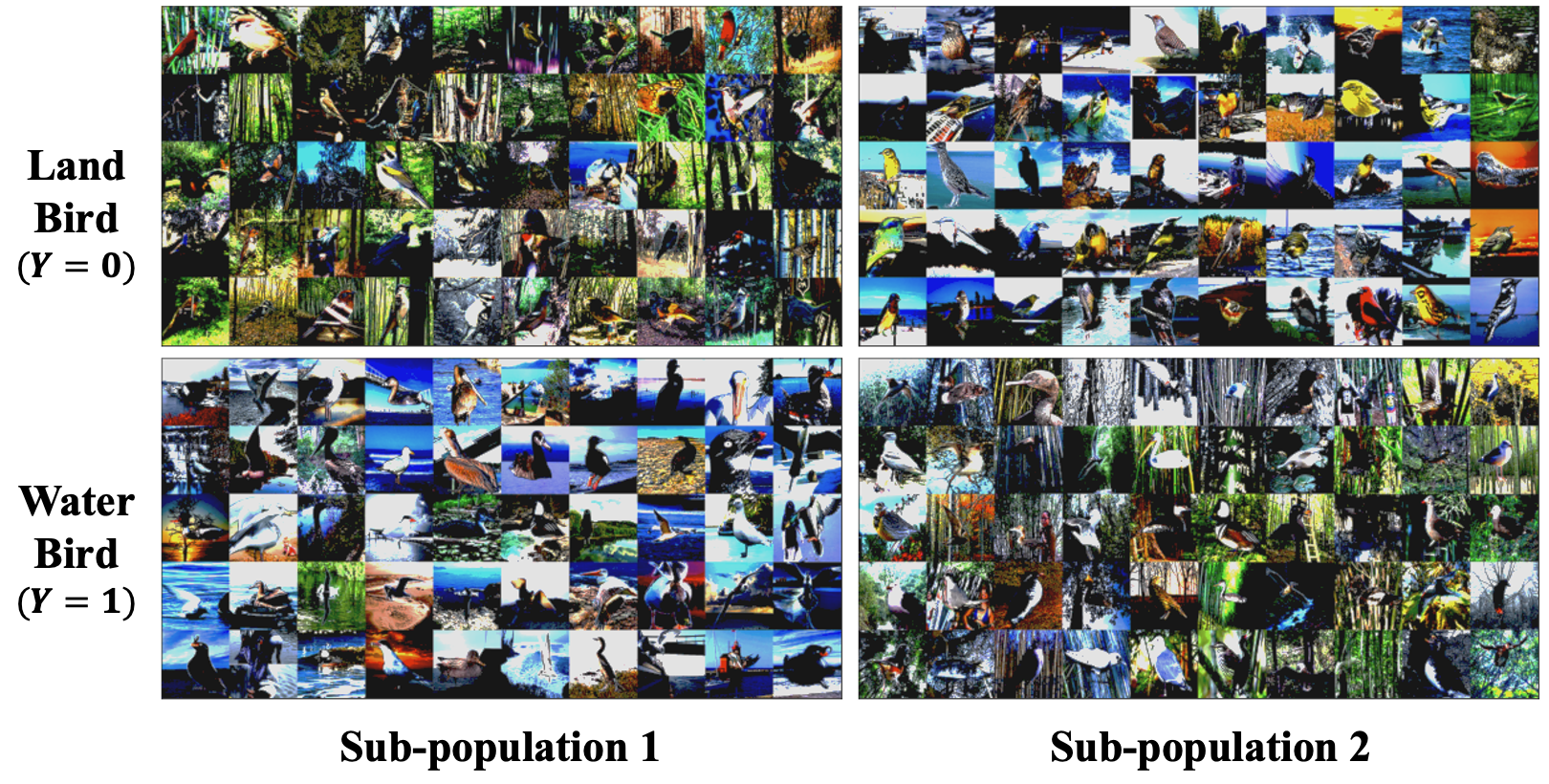}
  \caption{Results on the Waterbird data. Here we \emph{randomly sample} 50 images for each class and each learned sub-population.}
  \label{fig:test2}
\end{minipage}
\vskip -0.2in
\end{figure}

\subsection{Reveal Explainable Sub-population Structures}
The predictive heterogeneity could provide valuable insights for the sub-population division and support decision-making across various fields, including agricultural and sociological research, as well as object recognition.
Our illustrative examples below reveal that the learned sub-population divisions are highly explainable and relevant for decision-making purposes.

\textbf{Example: Agriculture}\quad  It is known that the climate affects crop yields and crop suitability \citep{lobell2008prioritizing}.
We utilize the data from the NOAA database which contains daily weather from weather stations around the world.
Following \cite{zhao2021comparing}, we extracted summary statistics from the weather sequence of the year 2018, including the average yearly temperature, humidity, wind speed and rainy days.
The task is to predict the \emph{crop yield} in each place with \emph{weather summary statistics} and \emph{location covariates (i.e. longitude and latitude)} of the place.
For easy illustration, we focus on the places with crop types of wheat or rice.
Notably, our input covariates do \emph{not} contain the crop type information. 
We use MLP models in this task and set $K=2$ for our IM algorithm.

Given that crop yield prediction mechanisms are closely related to crop type, which is unknown in the prediction task, we believe this causes data heterogeneity in the entire dataset, and the recognized predictive heterogeneity should relate to it. 
To demonstrate the rationality of our measure, we plot the real distribution map of wheat and rice planting areas in Figure \ref{fig:climate}(a) and the learned two sub-populations of our IM algorithm in Figure \ref{fig:climate}(b). 
The division given by our algorithm is quite similar to the real division of the two crops, indicating the rationality of our measure. 
We observe some discrepancies in areas such as the Tibet Plateau in Asia, which we attribute to the absence of significant features such as population density and altitude that significantly affect crop yields.

\textbf{Example: Sociology}\quad 
We use the UCI Adult dataset \citep{misc_adult_2}, which is widely used in the study of algorithmic fairness and derived from the 1994 Current Population Survey conducted by the US Census Bureau.
The task is to predict whether the income of a person is greater or less than 50k US dollars based on personal features.
We use linear models in this task and set $K=2$.
In this example, we aim to investigate whether \emph{sub-population structures} within data affect the learning of machine learning models.

In Figure \ref{fig:test1} (a), we plot summary statistics for the two sub-populations, revealing a key difference in capital gain.
In Figure \ref{fig:test1} (b), we present the feature importance given by linear models for the two sub-populations, and find that for individuals with high capital gain, the prediction model mainly relies on capital gain, which is fair.
However, for individuals with low capital gain, models also consider sensitive attributes such as sex and marital status, which have been known to cause discrimination.
Our results are consistent with those found in \citep{zhao2021comparing} and can help identify potential inequalities in decision-making.
For example, our findings suggest potential discrimination towards individuals with low capital gain, which could motivate algorithmic design and improve policy fairness.

\textbf{Example: Object Recognition}\quad Finally, we utilize the Waterbird dataset \citep{sagawa2019distributionally}, which is widely used as a benchmark in the field of robust learning, to investigate the impact of spurious correlations on machine learning models.
The task is to recognize waterbirds or landbirds, but the images contain \emph{spurious correlations} between the background and the target label. 
For the majority of images, waterbirds are located on water and landbirds on land, whereas for a minority of images, this correlation is reversed. 
Therefore, the spurious correlation leads to predictive heterogeneity in this dataset, which could significantly affect the performance of machine learning models.
In this example, we use the ResNet18 and set $K=2$ in our IM algorithm.

Our method successfully captures the spurious correlation and identifies two sub-populations of images with inverse correlations between the object and the background.
To demonstrate the effectiveness of our method, we randomly sample 50 images for each class and each learned sub-population and plot them in Figure \ref{fig:test2}. 
In sub-population 1, the majority of landbirds are on the ground and waterbirds are in the water, while in sub-population 2, the majority of landbirds are in the water and waterbirds are on the ground.
Our findings suggest that the proposed approach can be leveraged by robust learning methods \citep{sagawa2019distributionally, koyama2020out} to improve the generalization ability of machine learning models. 
By eliminating the influence of spurious correlations, our method could significantly enhance the robustness and reliability of machine learning models. 
Overall, our study highlights the importance of addressing predictive heterogeneity in image classification tasks and provides a practical solution for achieving robust learning performance.

\begin{figure}[htbp]
\centering
\begin{minipage}{.48\textwidth}
  \centering
  \includegraphics[width=\linewidth]{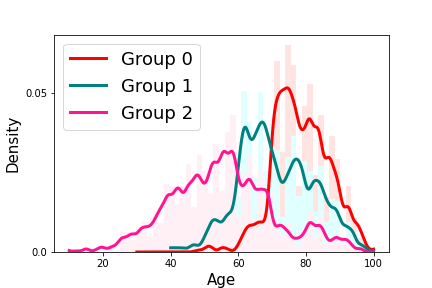}
  \caption{Results on the COVID-19 data. We plot the age distributions of dead people ($Y=1$) in each learned subgroup.}
  \label{fig:COVID}
\end{minipage}%
\hfill
\begin{minipage}{.48\textwidth}
  \centering
  \includegraphics[width=\linewidth]{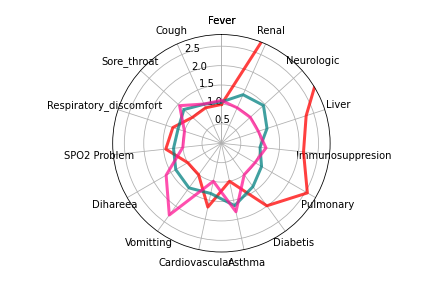}
  \caption{Results on the COVID-19 data. We show the averages of typical features of dead people ($Y=1$) in each learned subgroup.}
  \label{fig:COVID2}
\end{minipage}
\vskip -0.2in
\end{figure}

\subsection{Assist Scientific Discovery: Uncover Factors Related to Mortality}
In order to fully demonstrate the efficacy of our predictive heterogeneity, we focus on the application of healthcare, utilizing the COVID-19 dataset of Brazilian patients. 
This dataset comprises 6882 COVID-positive patients from Brazil, whose data was recorded between February 27th and May 4th, 2020. 
The dataset includes a wide range of risk factors, including comorbidities, symptoms, and demographic characteristics. 
The binary label corresponds to mortality caused by COVID-19. 
Our aim is to validate the sub-populations learned through our methodology on this dataset, by thoroughly \emph{explaining each group} and showcasing how our predictive heterogeneity can be employed to \emph{uncover features related to mortality that are otherwise difficult to detect}.

\subsubsection{Learned Sub-populations.}
When predicting mortality based on risk factors, it is important to consider that patients with various underlying diseases and demographic characteristics, such as age and sex, may exhibit different probabilities of mortality. 
Furthermore, it is plausible that the mortality of different individuals can be attributed to distinct factors. 
In light of these considerations, the predictive heterogeneity for this dataset is caused by the diversity of mechanisms that contribute to mortality among various sub-populations.

In this experiment, we use linear models and the loss function is binary cross-entropy loss. 
We select the sub-population number $K\in \{2,3,4,5,6\}$ that exhibit the maximal empirical predictive heterogeneity$\hat{\mathcal{H}}_{\mathcal V}^{\mathscr E_K}(X\rightarrow Y;\mathcal D)$, which results in three distinct subgroups (the optimal $K=3$).
Besides, we empirically observe that when $K>3$, the learned sub-populations will shrink to 3 sub-populations.
In Figure \ref{fig:COVID} and \ref{fig:COVID2}, to conduct a more thorough examination of the learned subgroups, we analyze the age distribution of each group, as well as the average value of their corresponding risk factors. 
Our analysis reveals several noteworthy findings:
\begin{itemize}
	\item[1.] We observe a distinct difference in the age distribution of the learned subgroups. Specifically, Group 0 is primarily composed of individuals over the age of 70, while Group 1 consists of individuals around 60 years old. 
Group 2, on the other hand, is comprised of middle-aged individuals spanning multiple age groups.
	\item[2.]  The average values of the risk factors reveal notable differences among the various subgroups, indicative of distinct causes of mortality. More specifically, Group 0 exhibits a considerably higher prevalence of underlying diseases, such as renal, neurologic, liver, and immunosuppression, when compared to the other groups. In contrast, Group 1 shows a substantially lower level of underlying diseases in comparison. Interestingly, Group 2 does not exhibit any underlying diseases, yet has a markedly higher level of diarrhea and vomiting. These findings suggest that the learned subgroups may be used to identify specific risk factors associated with mortality, which can inform targeted interventions for individuals with distinct risk profiles.
\end{itemize}
Having identified distinct patterns among the subgroups, we seek to identify the specific risk factors associated with mortality. 
To further validate our findings, we incorporate the expertise of domain experts. 
By leveraging their insights, we are able to confirm the reliability of the identified risk factors and the importance of our subgroup analysis.

\subsubsection{Scientific Findings}
Based on the learned group, we fit a logistic regression model on each group and pick the top-6 features with the largest coefficients, which are shown in Table \ref{tab:top-features}.

Firstly, our analysis reveals that in Group 0 and 1, the top features associated with mortality are primarily SPO2 and underlying diseases, which align with the common risk factors of older individuals. 
In contrast, Group 2 exhibits a distinct set of top features, including symptoms of COVID-19 such as fever, cough, and vomiting. 
Notably, Group 2 is composed of middle-aged individuals spanning multiple age groups. 
Our findings suggest that severe COVID-19 symptoms can lead to mortality regardless of age.

Secondly, to further our analysis, we fit a model for the entire dataset and observe that the top features remain SPO2 and underlying diseases, consistent with the top features found for older individuals. 
However, this may not be beneficial or could even lead to harm for interventions targeted towards younger or middle-aged individuals who generally do not have severe underlying diseases. 
For instance, doctors may tend to treat younger patients with severe COVID-19 symptoms optimistically and underestimate their mortality risk because they typically do not have underlying diseases.
Thus, exploring and leveraging the predictive heterogeneity within the data can lead to more reliable scientific discoveries while avoiding potential harm caused by latent heterogeneity.

Thirdly, our analysis reveals two important features in Group 2, namely vomiting and diarrhea, which are rarely considered in traditional analysis. 
We have reviewed relevant literature on COVID-19 and discovered that various studies have recognized these two symptoms as important indicators of higher risk of mortality caused by COVID-19. 
\citet{2020COVID} highlighted the potential mechanisms of gastrointestinal and hepatic injuries in COVID-19 to raise awareness of digestive system injury in COVID-19. 
\citet{Liu442} analyzed 29,393 laboratory-confirmed COVID-19 patients diagnosed before March 21, 2020, in cities outside of Wuhan in mainland China and found that patients with both GI symptoms and fever and patients with fever alone had a significantly higher risk of death, where GI symptoms refer to one of the following symptoms: nausea, vomiting, diarrhea, or abdominal pain. 
\citet{2021Gastrointestinal} also found that gastrointestinal symptoms are associated with the severity of COVID-19, and the severe rate was more than 40\% in COVID-19 patients with gastrointestinal symptoms. 
\citet{0Diarrhea} demonstrated that the presence of diarrhea as a presenting symptom is associated with increased disease severity and likely worse prognosis. 
\citet{2022COVID} have called for the consideration of COVID-19 in the differential diagnosis for patients who present with abdominal pain and gastrointestinal symptoms typical of gastroenteritis or surgical abdomen, even if they lack respiratory symptoms of COVID-19. 
These studies validate the reliability of our findings and demonstrate that studies utilizing the proposed predictive heterogeneity can uncover unusual risk factors that do not appear in analysis of the overall dataset.

This example serves as an illustration of the potential benefits that our predictive heterogeneity can offer to a wide range of scientific fields.
By exploiting the heterogeneity within a dataset, our approach can reveal novel patterns and relationships that may be overlooked in traditional analyses, leading to more reliable and comprehensive scientific discoveries

\begin{table}[htbp]
\centering
\caption{Top features of each learned subgroup and overall data on the COVID-19 dataset.}
\label{tab:top-features}
\resizebox{\textwidth}{1.2cm}{
\begin{tabular}{c|llllll}
\hline
Group ID & \multicolumn{6}{c}{Top Features} \\ \hline
0        &    SPO2 &   Diabetes  &   Renal  & Neurologic    &  Pulmonary   &  Cardiovascular  \\ \hline
1        &    Diabetes &  SPO2   &  Neurologic   & Cardiovascular    & Pulmonary     &  Renal  \\ \hline
2        &    \bf Fever & \bf Cough   & Renal    & \bf Vomitting    & \bf Shortness of breath    &  \bf Dihareea  \\ \hline
All      &    SPO2 &    Renal &   Neurologic  &   Diabetes  & Pulmonary    &  Cardiovascular  \\ \hline
\end{tabular}}
\end{table}

\subsection{Benefit Generalization}
\label{section:ood}

\begin{table}[b]
\vskip -0.1in
\caption{Results of the experiments on out-of-distribution generalization, including the simulated data and colored MNIST data.}
\label{table:results}
\centering\resizebox{\textwidth}{1.9cm}{
\begin{tabular}{|cc|cccc|||cc|}
\hline
\multicolumn{2}{|c|}{\multirow{3}{*}{\large Method}} & \multicolumn{4}{c|||}{\textbf{\large 1. Simulated Data}} & \multicolumn{2}{c|}{\textbf{\large 2. Colored MNIST}}\\ 
\multicolumn{2}{|c|}{}     &  \multicolumn{2}{c}{\bf Training Sub-population Error} & \multicolumn{2}{c|||}{\bf Test Error} & \multirow{2}{*}{\bf Train Accuracy} & \multicolumn{1}{c|}{\multirow{2}{*}{\bf Test Accuracy}}                                                                                                                  \\  
\multicolumn{2}{|l|}{}                                                                  & \multicolumn{1}{c}{Major ($r=1.9$)}            & \multicolumn{1}{c}{Minor ($r=-1.9$)}           & \multicolumn{1}{c}{$r=-2.3$}           & $r=-2.7$    &  &        \\ \hline
\multicolumn{2}{|c|}{ERM}                                 & \multicolumn{1}{c}{0.255{(\scriptsize$\pm 0.024$)}} & \multicolumn{1}{c|}{0.740{\scriptsize($\pm 0.022$)}} & \multicolumn{1}{c}{0.738{\scriptsize($\pm 0.035$)}} & 0.737{\scriptsize($\pm 0.023$)} & 0.998{\scriptsize($\pm 0.001$)} & 0.406{\scriptsize($\pm 0.019$)}  \\ 
\multicolumn{2}{|c|}{EIIL}                                   & \multicolumn{1}{c}{\bf 0.164{\scriptsize($\pm 0.014$)}} & \multicolumn{1}{c|}{1.428{\scriptsize($\pm 0.035$)}} & \multicolumn{1}{c}{1.431{\scriptsize($\pm 0.061$)}} & 1.431{\scriptsize($\pm 0.046$)} & 0.812{\scriptsize($\pm 0.006$)} & 0.610{\scriptsize($\pm 0.016$)}\\ \cline{1-2}
\multicolumn{1}{|c}{\multirow{3}{*}{\large KMeans}} & Balance   & \multicolumn{1}{c}{0.231{\scriptsize($\pm 0.022$)}} & \multicolumn{1}{c|}{0.847{\scriptsize($\pm 0.024$)}} & \multicolumn{1}{c}{0.846{\scriptsize($\pm 0.039$)}} & 0.845{\scriptsize($\pm 0.026$)}& \bf 0.999{\scriptsize($\pm 0.001$)} & 0.328{\scriptsize($\pm 0.021$)} \\ 
\multicolumn{1}{|c}{}                        & IRM        & \multicolumn{1}{c}{0.231{\scriptsize($\pm 0.022$)}} & \multicolumn{1}{c|}{0.845{\scriptsize($\pm 0.024$)}} & \multicolumn{1}{c}{0.844{\scriptsize($\pm 0.039$)}} & 0.843{\scriptsize($\pm 0.026$)} & 0.947{\scriptsize($\pm 0.004$)} & 0.259{\scriptsize($\pm 0.021$)}\\
\multicolumn{1}{|c}{}                        & IGA         & \multicolumn{1}{c}{0.235{\scriptsize($\pm 0.022$)}} & \multicolumn{1}{c|}{0.840{\scriptsize($\pm 0.023$)}} & \multicolumn{1}{c}{0.839{\scriptsize($\pm 0.038$)}} & 0.838{\scriptsize($\pm 0.027$)} & 0.997{\scriptsize($\pm 0.001$)} & 0.302{\scriptsize($\pm 0.021$)}\\ \cline{1-2}
\multicolumn{1}{|c}{\multirow{3}{*}{\large Ours}}   & Balance   & \multicolumn{1}{c}{0.403{\scriptsize($\pm 0.041$)}} & \multicolumn{1}{c|}{\bf 0.423{\scriptsize($\pm 0.016$)}} & \multicolumn{1}{c}{\bf 0.416{\scriptsize($\pm 0.022$)}} & \bf 0.416{\scriptsize($\pm 0.014$)} & 0.749{\scriptsize($\pm 0.012$)} & \bf 0.692{\scriptsize($\pm 0.039$)} \\ 
\multicolumn{1}{|c}{}                        & IRM        & \multicolumn{1}{c}{0.391{\scriptsize($\pm 0.039$)}} & \multicolumn{1}{c|}{\bf 0.432{\scriptsize($\pm 0.016$)}} & \multicolumn{1}{c}{\bf 0.430{\scriptsize($\pm 0.022$)}} &\bf 0.430{\scriptsize($\pm 0.014$)}  & 0.759{\scriptsize($\pm 0.014$)} & \bf 0.727{\scriptsize($\pm 0.047$)}\\
\multicolumn{1}{|c}{}                        & IGA        & \multicolumn{1}{c}{0.449{\scriptsize($\pm 0.037$)}} & \multicolumn{1}{c|}{\bf 0.426{\scriptsize($\pm 0.017$)}} & \multicolumn{1}{c}{\bf 0.417{\scriptsize($\pm 0.022$)}} &\bf 0.417{\scriptsize($\pm 0.014$)}  & 0.759{\scriptsize($\pm 0.012$)} & \bf  0.713{\scriptsize($\pm 0.034$)}\\ \hline
\end{tabular}
}
\end{table}


\begin{figure}
\centering
\begin{minipage}{1.0\textwidth}
  \centering
\subfigure[KMeans.] {
 \label{fig:a}     
\includegraphics[width=0.3\linewidth]{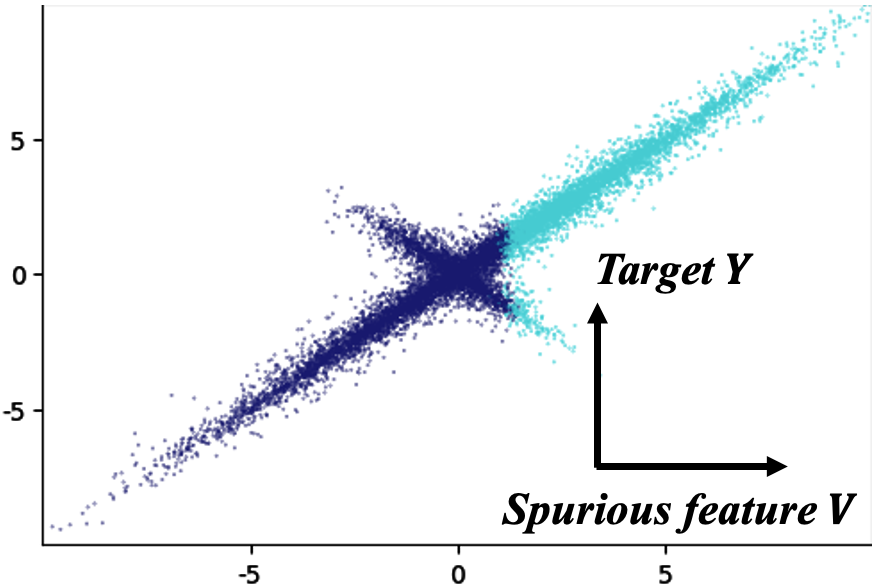}  
}      
\subfigure[EIIL.] {
 \label{fig:a}     
\includegraphics[width=0.3\linewidth]{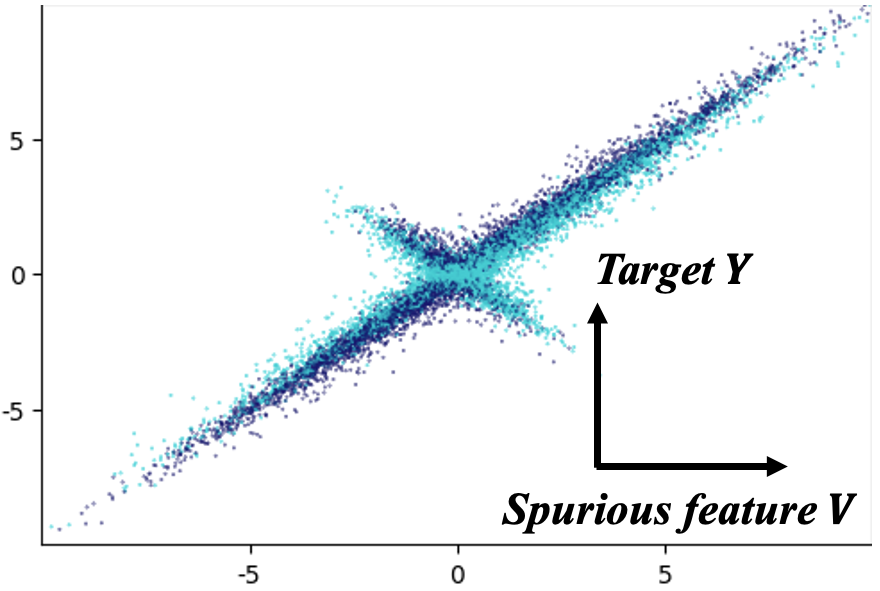}  
}   
\subfigure[Our IM.] {
 \label{fig:a}     
	\includegraphics[width=0.3\linewidth]{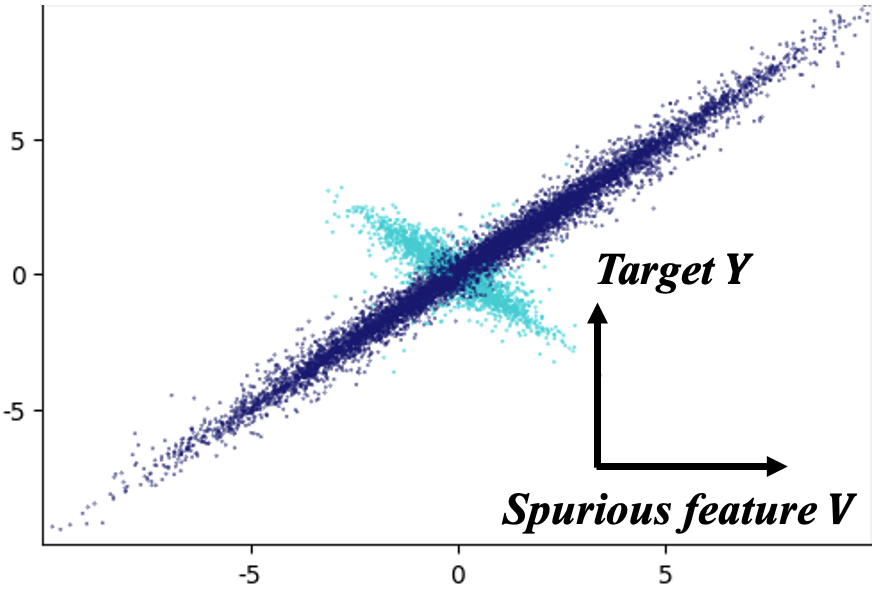}  
	}
	\vskip -0.1in
  \caption{Sub-population division on the simulated data of three methods, where two colors denote two sub-populations.}
  \label{fig:test3}
\end{minipage}
\end{figure}

 In this section, we aim to evaluate the efficacy of our IM algorithm in enhancing the out-of-distribution (OOD) generalization performance of machine learning models. 
 To this end, we conduct experiments on both simulated data and real-world colored MNIST data. 
 Our results suggest that the learned sub-population structures by our IM algorithm could significantly benefit the OOD generalization of machine learning models.

\textbf{Baselines}\quad First, we compare with \emph{empirical risk minimization} (ERM) and \emph{environment inference for invariant learning} (EIIL, \citep{creager2021environment}) which infers the environments for learning invariance.
Then we compare with the well-known \emph{KMeans} algorithm, which is the most popular clustering algorithm.
For our IM algorithm and KMeans, we involve three algorithms as backbones to leverage the learned sub-populations, including sub-population balancing and invariant learning methods.
The sub-population balancing simply equally weighs the learned sub-populations.
\emph{invariant risk minimization} (IRM, \citep{arjovsky2019invariant}) and \emph{inter-environment gradient alignment} (IGA, \citep{koyama2020out}) are typical methods in OOD generalization, which take the sub-populations as input environments to learn the invariant models.

\subsubsection{Simulation Data of Sample Selection Bias}
The input features $X=[S,T,V]^T\in\mathbb{R}^{10}$ consist of stable features $S\in\mathbb{R}^5$, noisy features $T\in\mathbb{R}^4$ and the spurious feature $V\in\mathbb{R}$:
\begin{small}
\begin{equation}
	S\sim\mathcal{N}(0,2\textbf{I}_5), T\sim\mathcal{N}(0,2\textbf{I}_4), Y=\theta_S^TS + h(S)+\mathcal{N}(0,0.5), V\sim\text{Laplace}(\text{sign}(r)\cdot Y, 1/(5\ln |r|))
\end{equation}	
\end{small}
where $\theta_S\in\mathbb{R}^5$ is the coefficient and $h(S)=S_1S_2S_3$ is the nonlinear term.
$|r|>1$ is a factor for each sub-population, and here the data heterogeneity is brought by the \emph{endogeneity with hidden variable} \citep{fan2014challenges}.
$V$ is the \emph{spurious feature} whose relationship with $Y$ is unstable across sub-populations and is controlled by the factor $r$.
Intuitively, $\text{sign}(r)$ controls whether the spurious correlation between $V$ and $Y$ is positive or negative. 
And $|r|$ controls the strength of the spurious correlation, i.e. the larger $|r|$ means the stronger spurious correlation.
In \emph{training}, we generate 10000 points, where the major group contains 80\% data with $r=1.9$ (i.e. strong \emph{positive} spurious correlation) and the minor group contains 20\% data with $r=-1.9$ (i.e. strong \emph{negative} spurious correlation).
In \emph{testing}, we test the performances of the two groups respectively, and we also set $r=-2.3$ and $r=-2.7$ to simulate stronger distributional shifts.
We use linear regression and set $K=2$ for all methods, and we report the mean-square errors (MSE) of all methods.

The results over 10 runs are shown in Table \ref{table:results}.
From the results in Table \ref{table:results}, for both the simulated and colored MNIST data, the two backbones with our IM algorithm achieve \emph{the best OOD generalization performances}.
Also, for the simulated data, the learned predictive heterogeneity enables backbone algorithms to equally treat the majority and minority inside data (i.e. low-performance gap between 'Major' and 'Minor'), and significantly benefits the OOD generalization.
Further, we plot the learned sub-populations of our IM algorithm in Figure \ref{fig:test3}.
From Figure \ref{fig:test3}, compared with KMeans and EIIL, our predictive heterogeneity exploits the spurious correlation between $V$ and $Y$, and enables the backbone algorithms to eliminate it.

%

\subsubsection{Simulation Data of Hidden Variables}
\quad\textcolor{black}{Also, we add one more experiment to show that (1) when the chosen $K$ is smaller than the ground-truth, the performances of our methods will drop but are still better than ERM (2) when the chosen $K$ is larger, the performances are not affected much.}

\begin{figure}[htbp]
\centering
\begin{minipage}{.55\textwidth}
  \centering
  \includegraphics[width=\linewidth]{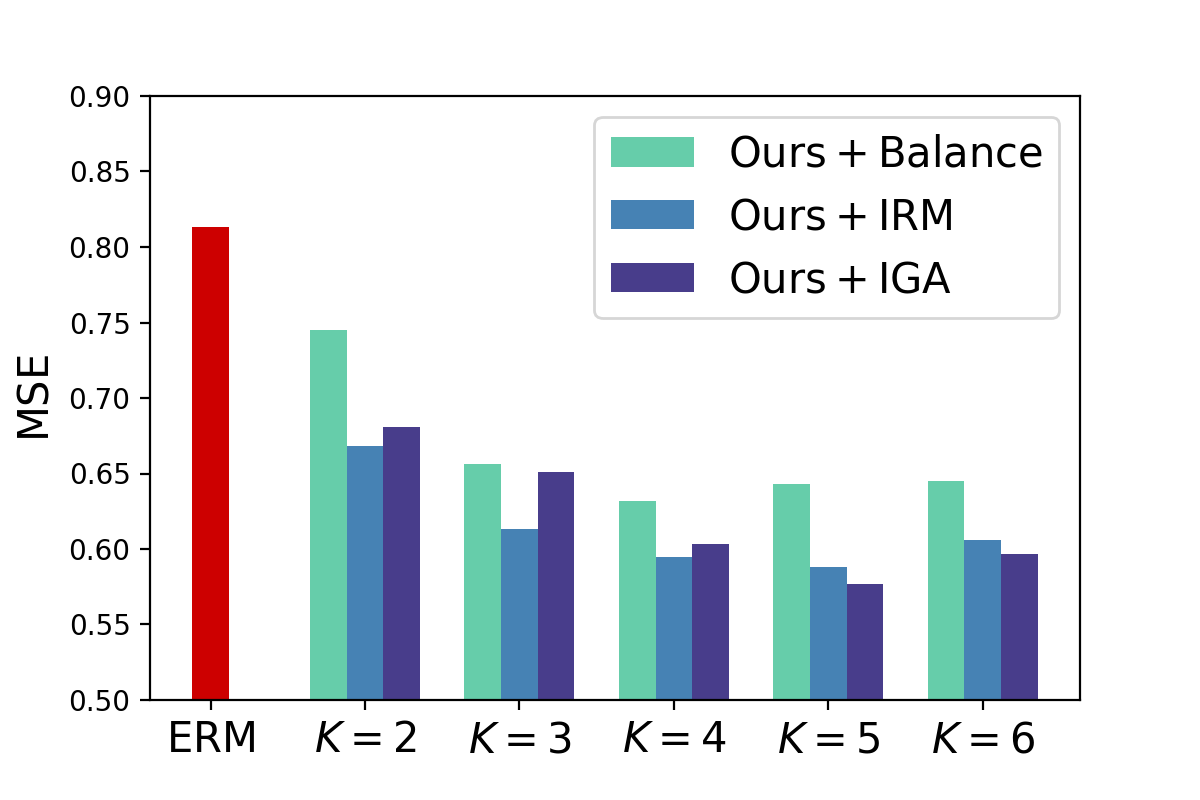}
  \caption{The OOD generalization error of our methods with Sub-population Balancing, IRM and IGA as backbones for the added experiments. The ground-truth sub-population number is 4.}
  \label{fig:appendix-rebuttal}
\end{minipage}%
\hfill
\begin{minipage}{.42\textwidth}
  \centering
  \includegraphics[width=\linewidth]{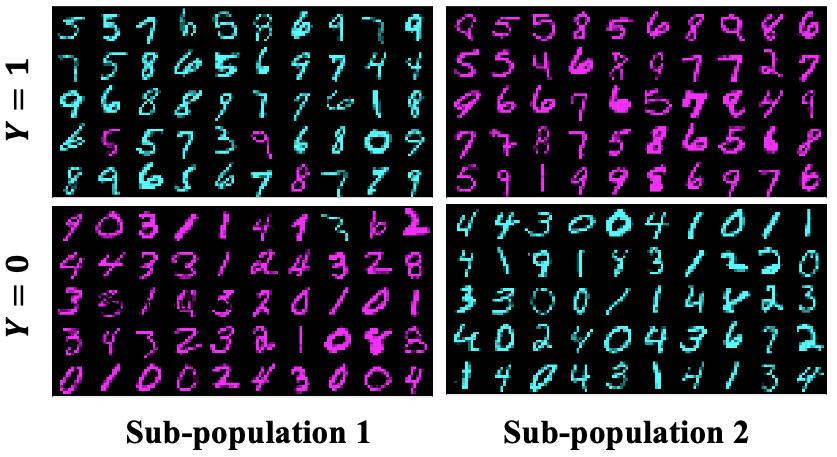}
  \caption{Sub-population division on the MNIST data of our IM algorithm.}
  \label{fig:test4}
\end{minipage}
\vskip -0.2in
\end{figure}

\textcolor{black}{The input features $X=[S,T,V]\in\mathbb{R}^{10}$ consist of stable features $S\in\mathbb{R}^5$, noisy features $T\in\mathbb{R}^4$ and the spurious feature $V\in\mathbb R$:
$$
S\sim \mathcal{N}(2,2\mathbb I_5),\quad  T\sim \mathcal{N}(0, 2\mathbb I_4), \quad Y=\theta_S^TS + S_1S_2S_3+\mathcal{N}(0,0.5),
$$
and we generate the spurious feature via:
$$
V = \theta_V^e Y + \mathcal{N}(0, 0.3),
$$
where $\theta_V^e$ varies across sub-populations and is dependent on which sub-population the data point belongs to.
In training, we sample 8000 data points from $e_1$ with $\theta_V^1=3.0$, 1000 points from $e_2$ with $\theta_V^2=-1.0$, 1000 points from $e_3$ with $\theta_V^3=-2.0$ and 1000 points from $e_4$ with $\theta_V^4=-3.0$.
Therefore, the ground-truth number of sub-populations is 4.
In testing, we test the performances on $e_4$ with $\theta_V^4=-3.0$, which has strong distributional shifts from training data.
The average MSE over 10 runs are shown in Figure \ref{fig:appendix-rebuttal}.}


\textcolor{black}{From the results, we can see that when $K$ is smaller than the ground-truth, increasing $K$ benefits the OOD generalization performance, and when $K$ is larger, the performances are not affected much.}

\textcolor{black}{For our IM algorithm, we think there are mainly two ways to choose $K$:}

\begin{itemize}
    \item \textcolor{black}{According to the predictive heterogeneity index: When the chosen $K$ is smaller than the ground-truth, our measure tends to increase quickly when increasing $K$; and when $K$ is larger than the ground-truth, the increasing speed will slow down, which could direct people to choose an appropriate $K$.}
    \item \textcolor{black}{According to the prediction model: Since our IM algorithm aims to learn sub-populations with different prediction mechanisms, one could compare the learned model parameters $\theta_1, \dots, \theta_K$ to judge whether $K$ is much larger than the ground-truth, i.e., if two resultant models are quite similar, $K$ may be too large (divide one sub-population into two). For linear models, one can directly compare the coefficients. For deep models, we think one can calculate the transfer losses across sub-populations.}
\end{itemize}


\subsubsection{Colored MNIST}
Following \cite{arjovsky2019invariant}, we design a binary classification task constructed on the MNIST dataset.
Firstly, digits $0\sim4$ are labeled $Y=0$ and digits $5\sim 9$ are labeled $Y=1$. 
Secondly, noisy labels $\tilde{Y}$ are induced by randomly flipping the label $Y$ with a probability of 0.2.
Then we sample the colored id $V$ spurious correlated with $\tilde{Y}$ as 
\begin{equation}
	V=\Big\{\begin{array}{ll}
     +\tilde{Y}, &\text{with probability }r,  \\
     -\tilde{Y}, &\text{with probability }1-r.
\end{array}
\end{equation}

In fact, $r$ controls the spurious correlation between $\tilde{Y}$ and $V$. 
In \emph{training}, we randomly sample 10000 data points and set $r=0.85$, meaning that for 85\% of the data, $V$ is positively correlated with $\tilde{Y}$ and for the rest 15\%, the spurious correlation becomes negative, which causes data heterogeneity w.r.t. $V$ and $\tilde{Y}$.
In \emph{testing}, we set $r=0$ (\emph{strong negative spurious correlation}), bringing strong shifts between training and testing.

From the results in Table \ref{table:results}, for both the simulated and colored MNIST data, the two backbones with our IM algorithm achieve \emph{the best OOD generalization performances}.
We plot the learned sub-populations of our IM algorithm in Figure \ref{fig:test4}.
From Figure \ref{fig:test4}, the learned sub-populations of our method also reflect the different directions of the spurious correlation between digit labels $Y$ and colors (red or green), which helps backbone methods to avoid using colors to predict digits.

\section{Related Work}
To the best of our knowledge, data heterogeneity has not converged to a uniform formulation so far, and has different meanings among different fields.
\cite{li1995definition} define the heterogeneity in \emph{ecology} based on the system property and complexity or variability.
\cite{rosenbaum2005heterogeneity} views the uncertainty of the potential outcome as unit heterogeneity in observational studies in \emph{economics}.
For \emph{graph} data, the heterogeneity refers to various types of nodes and edges (\cite{wang2019heterogeneous}).
More recently, in machine learning, several works of \emph{causal learning} \citep{peters2016causal, arjovsky2019invariant, koyama2020out, creager2021environment} and \emph{robust learning} \citep{sagawa2019distributionally} leverage heterogeneous data from multiple environments to improve the out-of-distribution generalization ability.
Specifically, invariant learning methods \citep{arjovsky2019invariant, koyama2020out, creager2021environment, zhou2022model} leverage the heterogeneous environment to learn the invariant predictors that have uniform performances across environments.
And in distributionally robust optimization field, \cite{sagawa2019distributionally, duchi2022distributionally} propose to optimize the worst-group prediction error to guarantee the OOD generalization performance.
However, in machine learning, previous works have not provided a precise definition or sound quantification of data heterogeneity, which makes it confusing and hard to leverage to develop more rational machine learning algorithms.

As for clustering algorithms, most algorithms only focus on the covariates $X$, typified by KMeans and Gaussian Mixture Model (GMM, \citep{reynolds2009gaussian}).
However, the learned clusters by KMeans can only reflect heterogeneous structures in $P(X)$, which is shown by our experiments.
Notably that our predictive heterogeneity could reflect the heterogeneity in $P(Y|X)$.
And the expectation maximization (EM, \citep{moon1996expectation}) can also be used for clustering.
However, our IM algorithm has essential differences from EM, for our IM algorithm infers latent variables that maximizes the predictive heterogeneity but EM maximizes the likelihood.
Also, there are methods \citep{creager2021environment} from the invariant learning field to infer environments.
Though it could benefit the OOD generalization, it lacks the theoretical foundation and only works in some settings.

\section{Discussion on differences with sub-group discovery}
\textcolor{black}{Subgroup discovery (SD, \citep{helal2016subgroup}) is aimed at extracting "interesting" relations among different variables ($X$) with respect to a target variable $Y$. Coverage and precision of each discovered group is the focus of such method. To be specific, it learns a partition on $P(X)$ such that some target label $y$ dominates within each group. The most siginficant gap between subgroup discovery and our predictive heterogeneity lies in the pattern of distributional shift among clusters: for subgroup discovery, $P(X)$ and $P(Y)$ varies across subgroups but there is a universal $P(Y|X)$. While for predictive heterogeneity $P(Y|X)$ differs across sub-population, which indicates diversified prediction mechanism. It is such disparity of prediction mechanism that inhibits   the performance of a universal predictive model on a heterogeneous dataset, which is the emphasis of OOD problem and group fairness.} 

\textcolor{black}{We think sub-group discovery is more applicable for settings where the distributional shift is minor while high explainability is required, since it generates simplified rules that people can understand. Also, sub-group discovery methods is suitable for the settings that only involve tabular data (typlically from a relational database), where the input features have clear semantics. 
And our proposed method could deal with general machine learning settings, including complicated data (e.g., image data) that involves representation learning.
Also, when people have to handle settings where data heterogeneity w.r.t. prediciton mechanism exists inside data, our method is more applicable.
However, both kinds of methods can be used to help people understand data and make more reasonable decisions.}

\section{Discussion on the Potential for fairness}
\textcolor{black}{We find combining our measure with algorithmic fairness is an interesting and promising direction and we think our measure has the potential to deal with algorithmic bias. 
Our method could generate sub-populations with possibly different prediction mechanisms, which could do some help in the following aspects:}

\textcolor{black}{\textbf{Risk feature selection}: we could select features according to our predictive heterogeneity measure to see what features bring the largest heterogeneity. If they are sensitive features, people should avoid their effects, and if they are not, they could direct people to build better machine learning models.}

\textcolor{black}{\textbf{Examine the algorithmic fairness}: we could use the learned sub-populations to examine whether a given algorithm is fair by calculating the performance gap across the sub-populations.}
\section{Conclusion}
We define the predictive heterogeneity, as the first quantitative formulation of the data heterogeneity that affects the prediction of machine learning models.
We demonstrate its theoretical properties and show that it benefits the out-of-distribution generalization performances.




\newpage

\appendix
\section{Proof of Proposition \ref{proposition1}}
\label{proof: prop1}
\begin{proof}[Proof of Proposition \ref{proposition1}]
\\\\
1. \emph{Monotonicity}:

Because of $\mathscr E_1 \subseteq \mathscr E_2$,
\begin{small}
\begin{align}
   \mathcal{H}^{\mathscr E_1}_{\mathcal V}(X \rightarrow Y) &= \sup_{\mathcal{E} \in \mathscr E_1}\mathbb{I}_{\mathcal{V}}(X\rightarrow Y|\mathcal{E})-\mathbb{I}_{\mathcal{V}}(X\rightarrow Y) \\
    &\leq \sup_{\mathcal{E} \in \mathscr E_2}\mathbb{I}_{\mathcal{V}}(X\rightarrow Y|\mathcal{E})-\mathbb{I}_{\mathcal{V}}(X\rightarrow Y) \\
    &= \mathcal{H}^{\mathscr E_2}_{\mathcal V}(X \rightarrow Y).
\end{align}
\end{small}
2. \emph{Nonnegativity}:

According to the definition of the environment set, there exists $\mathcal E_0 \in \mathscr E$ such that for any $e \in \text{supp}(\mathcal E)$, $X,Y|\mathcal E=e$ is identically distributed as $X,Y$. Thus, we have
\begin{small}
\begin{align}
    \mathcal{H}^{\mathscr E}_{\mathcal V}(X \rightarrow Y) &=
    \sup_{\mathcal{E} \in \mathscr E} \left[H_\mathcal V(Y|\emptyset,\mathcal E) - H_\mathcal V(Y|X,\mathcal E)\right] - \left[H_\mathcal V(Y|\emptyset) - H_\mathcal V(Y|X)\right] \\
    &\geq \left[H_\mathcal V(Y|\emptyset,\mathcal E_0) - H_\mathcal V(Y|X,\mathcal E_0)\right] - \left[H_\mathcal V(Y|\emptyset) - H_\mathcal V(Y|X)\right].
\end{align}
\end{small}
Specifically, 
\begin{small}
\begin{align}
    H_\mathcal V(Y|X,\mathcal E_0) &= \mathbb E_{e \sim \mathcal E_0} \left[ \inf\limits_{f\in\mathcal{V}}\mathbb{E}_{x,y\sim X,Y | \mathcal E=e}[-\log f[x](y)] \right] \\
    &= \mathbb E_{e \sim \mathcal E_0} \left[ \inf\limits_{f\in\mathcal{V}}\mathbb{E}_{x,y\sim X,Y}[-\log f[x](y)] \right] \\
    &= H_\mathcal V(Y|X).
\end{align}
\end{small}
Similarly, $H_\mathcal V(Y|\emptyset,\mathcal E_0) = H_\mathcal V(Y|\emptyset)$.
Thus, $\mathcal{H}^{\mathscr E}_{\mathcal V}(X \rightarrow Y) \geq 0$.\\\\
3. \emph{Boundedness}:

First, we have 
\begin{small}
\begin{align}
    H_{\mathcal V}(Y|X,\mathcal E) &= \mathbb E_{e \sim \mathcal E} \left[ \inf\limits_{f\in\mathcal{V}}\mathbb{E}_{x,y\sim X,Y|\mathcal E=e}[-\log f[x](y)] \right] \\
    &= \mathbb E_{e \sim \mathcal E} \left[ \inf\limits_{f\in\mathcal{V}}\mathbb{E}_{x\sim X|\mathcal E=e} \left[\mathbb E_{y \sim Y|x,e}[-\log f[x](y)] \right] \right]  \\
    &\geq 0,
\end{align}
\end{small}
by noticing that $\mathbb E_{y \sim Y|x}[-\log f[x](y)]$ is the cross entropy between $Y|x,e$ and $f[x]$.

Next,
\begin{small}
\begin{align}
    H_{\mathcal V}(Y|\emptyset,\mathcal E) &= \mathbb E_{e \sim \mathcal E} \left[ \inf\limits_{f\in\mathcal{V}}\mathbb{E}_{y\sim Y|\mathcal E=e}[-\log f[\emptyset](y)] \right] \\
    \label{equ:proposition1_3_1}
    &\leq \inf\limits_{f\in\mathcal{V}} \mathbb E_{e \sim \mathcal E} \left[ \mathbb{E}_{y\sim Y|\mathcal E=e}[-\log f[\emptyset](y)] \right] \\
    &= \inf\limits_{f\in\mathcal{V}} \mathbb{E}_{y\sim Y}[-\log f[\emptyset](y)] \\
    &= H_{\mathcal V}(Y|\emptyset),
\end{align}
\end{small}
where Equation \ref{equ:proposition1_3_1} is due to Jensen's inequality.

Combing the above inequalities,
\begin{align}
    \mathcal{H}^{\mathscr E}_{\mathcal V}(X \rightarrow Y) &=
    \sup_{\mathcal{E} \in \mathscr E} \left[H_\mathcal V(Y|\emptyset,\mathcal E) - H_\mathcal V(Y|X,\mathcal E)\right] - \left[H_\mathcal V(Y|\emptyset) - H_\mathcal V(Y|X)\right] \\
    &\leq  \sup_{\mathcal{E} \in \mathscr E} H_\mathcal V(Y|\emptyset,\mathcal E)  - \left[H_\mathcal V(Y|\emptyset) - H_\mathcal V(Y|X)\right] \\
    &\leq H_\mathcal V(Y|\emptyset) - \left[H_\mathcal V(Y|\emptyset) - H_\mathcal V(Y|X)\right] \\
    &= H_\mathcal V(Y|X).
\end{align}
4. \emph{Corner Case}:

According to Proposition 2 in \cite{DBLP:conf/iclr/XuZSSE20}, 
\begin{align}
    H_\Omega(Y|\emptyset) &= H(Y). \\
    H_\Omega(Y|X) &= H(Y|X).
\end{align}
By taking random variables $R,S$ identically distributed as $X,Y|\mathcal E=e$ for $e \in \text{supp}(\mathcal E)$, we have
\begin{align}
    H_{\Omega}(Y|X,\mathcal E=e) = H_{\Omega}(S|R) = H(S|R) = H(Y|X,\mathcal E=e).
\end{align}
Thus, 
\begin{align}
    H_\Omega(Y|X,\mathcal E) = \mathbb E_{e\sim \mathcal E}[H_\Omega(Y|X,\mathcal E=e)] = \mathbb E_{e\sim \mathcal E}[H(Y|X,\mathcal E=e)] = H(Y|X,\mathcal E).
\end{align}
Similarly, we have $ H_\Omega(Y|\emptyset,\mathcal E) = H(Y|\mathcal E)$.
Thus,
\begin{align}
    \mathcal{H}^{\mathscr E}_{\Omega}(X \rightarrow Y) &= 
    \sup_{\mathcal{E} \in \mathscr E} \left[H_\Omega(Y|\emptyset,\mathcal E) - H_\Omega(Y|X,\mathcal E)\right] - \left[H_\Omega(Y|\emptyset) - H_\Omega(Y|X)\right] \\
    &=  \sup_{\mathcal{E} \in \mathscr E} \left[H(Y|\mathcal E) - H(Y|X,\mathcal E)\right] - \left[H(Y) - H(Y|X)\right] \\
    &= \sup_{\mathcal{E} \in \mathscr E}\mathbb{I}(Y;X|\mathcal{E})-\mathbb{I}(Y;X) \\
    &= \mathcal{H}^{\mathscr E}(X, Y).
\end{align}
\end{proof}

\section{Proof of Theorem \ref{theorem: homogeneous}}
\label{proof:homogeneous}
\begin{proof}[Proof of Theorem \ref{theorem: homogeneous}]
\quad

1)

\begin{align}
    H_{\mathcal V_{\mathcal G}}(Y|X) &= \inf\limits_{f\in \mathcal V_{\mathcal G}}\mathbb{E}_{x\sim X} \left[\mathbb E_{y \sim Y|x}[-\log f[x](y)] \right] \\
    \label{equ:theorem1_1_1}
    &\leq \mathbb{E}_{x\sim X} \left[\mathbb E_{y \sim Y|x}[-\log \frac{1}{\sqrt{2\pi} \cdot \frac{1}{\sqrt{2\pi}}} \exp{ \left[-\frac{(y-g(x))^2}{2\cdot \frac{1}{2\pi} } \right] } \right] \\
    &= \mathbb{E}_{x\sim X} \left[\mathbb E_{y \sim Y|x}[\pi (y-g(x))^2 ] \right] = \pi\sigma^2.
\end{align}
Equation \ref{equ:theorem1_1_1} holds by taking $f[x] = \mathcal N(g(x), \frac{1}{2\pi})$.

2) 

Given the function family $\mathcal{V}_\sigma=\{f | f[x]=\mathcal{N}(\theta x,\sigma^2), \theta \in \mathbb R, \sigma \text{ fixed }\}$, by expanding the Gaussian probability density function in the definition of predictive $\mathcal V$-information, it could be shown that 
\begin{align}
\label{equ:Iv(X->Y)}
    \mathbb{I}_{\mathcal{V}_\sigma}(X\rightarrow Y) &\propto \min_{k\in \mathbb R} -\mathbb{E}[(Y-kX)^2] + \text{Var}(Y),
\end{align}
where the predictive $\mathcal V$-information is proportional to Mean Square Error subtracted by the variance of target, by a coefficient completely dependent on $\sigma$.

The minimization problem is solved by 
\begin{equation}
    k = \frac{\mathbb E[XY]}{\mathbb E[X^2]} = 1.
\end{equation}
Substituting $k$ into eq.\ref{equ:Iv(X->Y)},
\begin{align}
    \mathbb{I}_{\mathcal{V}_\sigma}(X\rightarrow Y) &\propto (-\mathbb E[\epsilon^2] + \text{Var}(X+\epsilon))=\text{Var}(X) = \mathbb E[X^2].
\end{align}
Denote $\text{supp}(\mathcal{E})=\{\mathcal{E}_1,\mathcal{E}_2\}$. Let $Q$ be the joint distribution of $(X,\epsilon,\mathcal E)$. 
Let $Q(\mathcal E_1)=\alpha$ and $ Q(\mathcal E_2)=1-\alpha $ be the marginal of $\mathcal E$. Abbreviate $Q(X,\epsilon|\mathcal E=\mathcal E_1)$ by $P_1(X,\epsilon)$ and $Q(X,\epsilon|\mathcal E=\mathcal E_2)$ by $P_2(X,\epsilon)$.

Similar to \ref{equ:Iv(X->Y)},
\begin{align}
\label{equ:Iv(X->Y|e)}
    \mathbb{I}_{\mathcal{V}_\sigma}(X\rightarrow Y|\mathcal E) &\propto \min_k -\mathbb{E}[(Y-kX)^2|\mathcal E] + \text{Var}(Y|\mathcal E).
\end{align}
For $\mathcal E=\mathcal E_1$, the minimization problem is solved by
\begin{align}
    k = \frac{\mathbb E_{P_1}[XY]}{\mathbb E_{P_1}[X^2]}.
\end{align}
Thus,
\begin{small}
\begin{align}
    \mathbb{I}_{\mathcal{V}_\sigma}(X\rightarrow Y|\mathcal E=\mathcal E_1) &\propto -\mathbb E_{P_1}\left[\left(Y-\frac{\mathbb E_{P_1}[XY]}{\mathbb E_{P_1}[X^2]}X\right)^2\right] + \text{Var}_{P_1}(Y) \\
    &= -\mathbb E_{P_1}[Y^2] + \frac{\mathbb E_{P_1}^2[XY]}{\mathbb E_{P_1}[X^2]} + (\mathbb E_{P_1}[Y^2] - \mathbb E_{P_1}^2[Y]) 
    = -\mathbb E_{P_1}^2[Y] + \frac{\mathbb E_{P_1}^2[XY]}{\mathbb E_{P_1}[X^2]}.
\end{align}
\end{small}
Similarly, we have
\begin{align}
\label{equ:Iv(X->Y|e_2)}
    \mathbb{I}_{\mathcal{V}_\sigma}(X\rightarrow Y|\mathcal E=\mathcal E_2) &\propto -\mathbb E_{P_2}^2[Y] + \frac{\mathbb E_{P_2}^2[XY]}{\mathbb E_{P_2}[X^2]}.
\end{align}
Notably, $\mathbb E_{P_1}[X^2]$ and $\mathbb E_{P_2}[X^2]$ are constrained by $\alpha$ and $\mathbb E[X^2]$.
\begin{equation}
    \mathbb E[X^2] = \mathbb E[\mathbb E[X^2|\mathcal E]] = \alpha \mathbb E_{P_1}[X^2] + (1-\alpha)\mathbb E_{P_2}[X^2].
\end{equation}
Similarly,
\begin{equation}
    \mathbb E[X^2] = \mathbb E[XY] = \alpha \mathbb E_{P_1}[XY] + (1-\alpha)\mathbb E_{P_2}[XY].
\end{equation}
\begin{equation}
    0 =\mathbb E[Y] = \alpha \mathbb E_{P_1}[Y] + (1-\alpha)\mathbb E_{P_2}[Y].
\end{equation}
The moments of $P_2$ could thereafter be represented by those of $P_1$.
\begin{small}
\begin{equation}
    \mathbb E_{P_2}[X^2] = \frac{\mathbb E[X^2] - \alpha \mathbb E_{P_1}[X^2]}{1-\alpha},
    \mathbb E_{P_2}[XY] = \frac{\mathbb E[X^2] - \alpha \mathbb E_{P_1}[XY]}{1-\alpha},
    \mathbb  E_{P_2}[Y] = - \frac{\alpha \mathbb E_{P_1}[Y]}{1-\alpha}.
\end{equation}
\end{small}
Substituting to eq.\ref{equ:Iv(X->Y|e_2)},
\begin{align}
    \mathbb{I}_{\mathcal{V}_\sigma}(X\rightarrow Y|\mathcal E=\mathcal E_2) &\propto -\frac{\alpha^2}{(1-\alpha)^2}E_{P_1}^2[Y] + \frac{1}{1-\alpha}\frac{\left(\mathbb E[X^2] - \alpha \mathbb E_{P_1}[XY]\right)^2}{\mathbb E[X^2] - \alpha \mathbb E_{P_1}[X^2]}.
\end{align}
Thus,
\begin{small}
\begin{align}
    \mathcal H_{\mathcal V_\sigma}^{\mathscr E}(X \rightarrow Y) &= \sup_{\mathcal E \in \mathscr E} -\mathbb{I}_{\mathcal{V}_\sigma}(X\rightarrow Y) + \alpha \mathbb{I}_{\mathcal{V}_\sigma}(X\rightarrow Y|\mathcal E=\mathcal E_1) + (1-\alpha) \mathbb{I}_{\mathcal{V}_\sigma}(X\rightarrow Y|\mathcal E=\mathcal E_2) \\
    &\propto \sup_{\mathcal E \in \mathscr E} -\mathbb E[X^2] - \alpha \mathbb E_{P_1}^2[Y] + \alpha \frac{\mathbb E_{P_1}^2[XY]}{\mathbb E_{P_1}[X^2]} - \frac{\alpha^2}{1-\alpha} \mathbb E_{P_1}^2[Y] + \frac{\left(\mathbb E[X^2] - \alpha \mathbb E_{P_1}[XY]\right)^2}{\mathbb E[X^2] - \alpha \mathbb E_{P_1}[X^2]} \\
    &= \sup_{\mathcal E \in \mathscr E} -\frac{\alpha}{1-\alpha}\mathbb E_{P_1}^2[X+\epsilon] + \alpha \frac{\mathbb E_{P_1}^2[X\epsilon]}{\mathbb E_{P_1}[X^2]\left(\mathbb E[X^2] - \alpha \mathbb E_{P_1}[X^2]\right)} \mathbb E[X^2].
\end{align}
\end{small}
Assuming $X \perp \epsilon \;|\; \mathcal E$, 
\begin{align}
    \mathcal H_{\mathcal V_\sigma}^{\mathscr E}(X \rightarrow Y) \propto \sup_{\mathcal E \in \mathscr E} -\frac{\alpha}{1-\alpha}\mathbb E_{P_1}^2[X+\epsilon] \leq 0.
\end{align}
From Proposition \ref{proposition1}, we have $\mathcal H_{\mathcal V_\sigma}^{\mathscr E}(X \rightarrow Y) \geq 0$. Thus, $\mathcal H_{\mathcal V_\sigma}^{\mathscr E}(X \rightarrow Y) = 0$.
\end{proof}

\section{Proof of Linear Cases (Theorem \ref{theorem: selection-bias} and \ref{theorem: omitted variable})}
\label{proof: linear}
\begin{proof}[Proof of Theorem \ref{theorem: selection-bias}]

For the ease of notion, we denote the $r(\mathcal{E}^*)$ as $r_e$, $\sigma(\mathcal{E}^*)$ as $\sigma_e$, and $\sigma(\mathcal{E}^*)\cdot\epsilon_v$ as $\epsilon_e$. 
And we omit the superscript $\mathcal C$ of $\mathcal{H}_{\mathcal V}^{\mathcal C}$.
	Firstly, we calculate the $H_{\mathcal{V}}[Y|\emptyset]$ as:
	\begin{align}
		H_{\mathcal{V}}[Y|\emptyset] &= \frac{1}{2\sigma^2}\text{Var}(Y) + \log\sigma + \frac{1}{2}\log 2\pi,\\
		H_{\mathcal{V}}[Y|\emptyset,\mathcal{E}^*] &= \frac{1}{2\sigma^2} \mathbb{E}_{\mathcal{E}^*}[\text{Var}(Y|\mathcal{E}^*)]+ \log\sigma + \frac{1}{2}\log 2\pi.
	\end{align}
	Therefore, we have
	\begin{equation}
		H_{\mathcal{V}}[Y|\emptyset,\mathcal{E}^*] - H_{\mathcal{V}}[Y|\emptyset] = -\frac{1}{2\sigma^2}\text{Var}(\mathbb{E}[Y|\mathcal{E}^*])\leq 0.
	\end{equation}
	As for $H_{\mathcal{V}}[Y|X]$, we have
	\begin{align}
		H_{\mathcal{V}}[Y|X] &= \inf_{h_S,h_V}\mathbb{E}_{X,Y}\left[\|Y-(h_SS+h_VV)\|^2\right]\frac{1}{2\sigma^2}\\
		&= \inf_{h_S,h_V}\mathbb{E}_{\mathcal{E}^*}\left[\mathbb{E}[\|f(S)+\epsilon_Y-(h_SS+h_V(r_ef(S)+\epsilon_e))\|^2|\mathcal{E}^*]\right]\frac{1}{2\sigma^2},
	\end{align}
	where we let $h_S=h_S-\beta$ here.
	Then we have
	\begin{align}
	2\sigma^2 H_{\mathcal{V}}[Y|X] &= \inf_{h_S,h_V}\mathbb{E}_{\mathcal{E}^*}\left[\mathbb{E}[\|(1-h_Vr_e)f(S)+\epsilon_Y-h_SS-h_V\epsilon_e\|^2|\mathcal{E}^*]\right]\\
		&= \inf_{h_S,h_V}\mathbb{E}_{\mathcal{E}^*}\left[\mathbb{E}[\|(1-h_Vr_e)f(S)-h_SS\|^2|\mathcal{E}^*]\right] + \sigma_Y^2 + h_V^2\mathbb{E}_{\mathcal{E}^*}[\sigma_e^2],
	\end{align}
	notably that here for $e_i,e_j\in\text{supp}(\mathcal{E}^*)$, we assume $P^{e_i}(S,Y)=P^{e_j}(S,Y)$ (we choose such $\mathcal{E}^*$ as one possible split).
	And the solution of $h_S,h_V$ is
	\begin{align}
		h_S &= \frac{\text{Var}(r_e)\mathbb{E}[f^2(S)]\mathbb{E}[f(S)S]+\mathbb{E}[\sigma_e^2]\mathbb{E}[f(S)S]}{\mathbb{E}[r_e^2]\mathbb{E}[f^2(S)]\mathbb{E}[S^2] + \mathbb{E}[\sigma_e^2]\mathbb{E}[S^2] - \mathbb{E}^2[r_e]\mathbb{E}^2[f(S)S]},\\
		h_V &= \frac{\mathbb{E}[r_e](\mathbb{E}[f^2(S)]\mathbb{E}[S^2]-\mathbb{E}^2[f(S)S])}{\mathbb{E}[r_e^2]\mathbb{E}[f^2(S)]\mathbb{E}[S^2] + \mathbb{E}[\sigma_e^2]\mathbb{E}[S^2] - \mathbb{E}^2[r_e]\mathbb{E}^2[f(S)S]}.
	\end{align}
	According to the assumption that $\mathbb{E}[f(S)S]=0$, we have
	\begin{align}
		h_S = 0,\quad
		h_V = \frac{\mathbb{E}[r(\mathcal E^*)]\mathbb{E}[f^2]}{\mathbb{E}[r^2(\mathcal E^*)]\mathbb{E}[f^2]+\mathbb{E}[\sigma^2(\mathcal E^*)]}.
	\end{align}
	Therefore, we have
	\begin{align}
		2\sigma^2 H_{\mathcal{V}}[Y|X] &= \mathbb{E}_{\mathcal{E}^*}[\mathbb{E}[\|(1-h_Vr_e)f(S)\|^2|\mathcal{E}^*]] + \sigma_Y^2+h_V^2\mathbb{E}_{\mathcal{E}^*}[\sigma_e^2]\\
		&= \frac{\text{Var}(r_e)\mathbb{E}[f^2]+\mathbb{E}[\sigma^2(\mathcal E^*)]}{\mathbb{E}[r_e^2]\mathbb{E}[f^2]+\mathbb{E}[\sigma^2(\mathcal E^*)]}\mathbb{E}[f^2(S)]+ \sigma_Y^2,\\
		2\sigma^2 H_{\mathcal{V}}[Y|X,\mathcal{E}^*] &= \sigma_Y^2+ \mathbb{E}[(\frac{1}{\frac{r_e^2\mathbb{E}[f^2]}{\sigma_e^2}+1})^2]\mathbb{E}[f^2]+ \mathbb{E}_{\mathcal{E}^*}[(\frac{1}{\frac{r_e}{\sigma_e}+\frac{\sigma_e}{r_e\mathbb{E}[f^2]}})^2].
	\end{align}
	Note that here we simply set $\sigma=1$ in the main body.
    And we have:
    \begin{equation}
        \mathcal{H}_{\mathcal V}(X\rightarrow Y)\approx \frac{\text{Var}(r_e)\mathbb{E}[f^2]+\mathbb{E}[\sigma^2(\mathcal E^*)]}{\mathbb{E}[r_e^2]\mathbb{E}[f^2]+\mathbb{E}[\sigma^2(\mathcal E^*)]}\mathbb{E}[f^2(S)]
    \end{equation}
    The approximation error is bounded by $\frac{1}{2}\max(\sigma_Y^2, R(r(\mathcal E^*), \sigma(\mathcal E^*), \mathbb{E}[f^2]))$, and $R(r(\mathcal E^*), \sigma(\mathcal E^*), \mathbb{E}[f^2])$ is defined as:
    \begin{equation}
        R(r(\mathcal E^*), \sigma(\mathcal E^*), \mathbb{E}[f^2]) = \mathbb{E}[(\frac{1}{\frac{r_e^2\mathbb{E}[f^2]}{\sigma_e^2}+1})^2]\mathbb{E}[f^2]+ \mathbb{E}_{\mathcal{E}^*}[(\frac{1}{\frac{r_e}{\sigma_e}+\frac{\sigma_e}{r_e\mathbb{E}[f^2]}})^2]
    \end{equation}
\end{proof}

\begin{proof}[Proof of Theorem \ref{theorem: omitted variable}]
	Similar as the above proof.	
\end{proof}

\section{Proof of the Error Bound for Finite Sample Estimation (Theorem \ref{theorem:pac})}
\label{proof: pac}

In this section, we will prove the error bound of estimating the predictive heterogeneity with the empirical predictive heterogeneity. Before the proof of Theorem \ref{theorem:pac} which is inspired by \cite{DBLP:conf/iclr/XuZSSE20}, we will introduce three lemmas.

\begin{lemma}
\label{lemma:err_1}
Assume $\forall x \in \mathcal X$,$\forall y \in \mathcal Y$,$\forall f \in \mathcal V$, $\log f[x](y) \in [-B,B]$ where $B > 0$. Define a function class $\mathcal G_{\mathcal V}^k = \{g|g(x,y) = \log f[x](y)q(\mathcal E=e_k|x,y), f\in \mathcal V, q \in \mathcal Q  \}$. Denote the Rademacher complexity of $\mathcal G$ with $N$ samples by $\mathscr R_{N}(\mathcal G)$. Define 
\begin{equation}
\hat f_k = \arg \inf_f \frac{1}{|\mathcal D|}  \sum_{x_i,y_i \in \mathcal D} -\log f[x_i](y_i) q(\mathcal E=e_k|x_i,y_i).
\end{equation}

Then for any $q \in \mathcal Q$,  any $\delta \in (0,1)$, with a probability over $1 - \delta$,  we have
\begin{small}
\begin{align}
    &\quad\; \left|q(\mathcal E=e_k)H_{\mathcal V}(Y|X,\mathcal E=e_k)  - \frac{1}{|\mathcal D|} \sum_{x_i,y_i \in \mathcal D} -\log \hat{f_k}[x_i](y_i) q(\mathcal E=e_k|x_i,y_i) \right| \\
    &\leq 2\mathscr R_{|\mathcal D|}(\mathcal G_{\mathcal V}^k) + B\sqrt{\frac{2\log{\frac{1}{\delta}}}{|\mathcal D|}}.
\end{align}
\end{small}
\end{lemma}
\begin{proof}
Apply McDiarmid's inequality to the function $\Phi(\mathcal D)$ which is defined as:
\begin{small}
\begin{align}
    \Phi(\mathcal D)  
    &= \sup_{f\in \mathcal V, q \in \mathcal Q} \left| q(\mathcal E=e_k)\mathbb E_{q} \left[ -\log f[x](y)|\mathcal E=e_k \right]  - \frac{1}{|\mathcal D|} \sum_{x_i,y_i \in \mathcal D} -\log f[x_i](y_i) q(\mathcal E=e_k|x_i,y_i)   \right|.
\end{align}
\end{small}
Let $\mathcal D$ and $\mathcal D'$ be two identical datasets except for one data point $x_j \neq x_j'$. We have:
\begin{small}
\begin{align}
    & \quad\; \Phi(\mathcal D) - \Phi(\mathcal D') \\
    & \leq \sup_{f\in \mathcal V, q \in \mathcal Q} \left[ \left| q(\mathcal E=e_k)\mathbb E_{q} \left[ -\log f[x](y)|\mathcal E=e_k \right]  - \frac{1}{|\mathcal D|} \sum_{x_i,y_i \in \mathcal D} -\log f[x_i](y_i) q(\mathcal E=e_k|x_i,y_i)   \right| \right.\\ 
    &\left. \quad\quad\quad\quad - \left| q(\mathcal E=e_k)\mathbb E_{q} \left[ -\log f[x](y)|\mathcal E=e_k \right]  - \frac{1}{|\mathcal D'|} \sum_{x_i',y_i' \in \mathcal D'} -\log f[x_i'](y_i') q(\mathcal E=e_k|x_i',y_i')   \right|\right] \\
    &\leq \sup_{f\in \mathcal V, q \in \mathcal Q} \left| \frac{1}{|\mathcal D|} \sum_{x_i,y_i \in \mathcal D} -\log f[x_i](y_i) q(\mathcal E=e_k|x_i,y_i) - \frac{1}{|\mathcal D'|} \sum_{x_i',y_i' \in \mathcal D'} -\log f[x_i'](y_i') q(\mathcal E=e_k|x_i',y_i')  \right| \\
    &=  \sup_{f\in \mathcal V, q \in \mathcal Q}\frac{1}{|\mathcal D|}  \left| \log f[x_j](y_j) q(\mathcal E=e_k|x_j,y_j) - \log f[x_j'](y_j') q(\mathcal E=e_k|x_j',y_j') \right| \\
    &\leq \frac{2B}{|\mathcal D|}.
\end{align}
\end{small}
According to McDiarmid's inequality, for any $\delta \in (0,1)$, with a probability over $1 - \delta$, we have:
\begin{align}
    \label{equ:err_7}
    \Phi(\mathcal D) \leq \mathbb E_{\mathcal D} [\Phi(\mathcal D)] + B\sqrt{\frac{2\log{\frac{1}{\delta}}}{|\mathcal D|}}.
\end{align}
Next we derive a bound for $\mathbb E_{\mathcal D}[\Phi(\mathcal D)]$.
Consider a dataset $\mathcal D'$ independently and identically drawn from $q(X,Y) = P(X,Y)$ with the same size as $\mathcal D$. We notice that
\begin{small}
\begin{align}
    q(\mathcal E=e_k)\mathbb E_{q} \left[ -\log f[x](y)|\mathcal E=e_k \right]
    = \mathbb E_{\mathcal D'} \left[ -\frac{1}{|\mathcal D'|} \sum_{x_i',y_i' \in \mathcal D'} -\log f[x_i'](y_i') q(\mathcal E=e_k|x_i',y_i') \right].
\end{align}
\end{small}
Thus, $\mathbb E_{\mathcal D}[\Phi(\mathcal D)]$ could be reformulated as:
\begin{small}
\begin{align}
    \mathbb E_{\mathcal D}[\Phi(\mathcal D)] &= \mathbb E_{\mathcal D}\left[ \sup_{f\in \mathcal V, q \in \mathcal Q}  \left| \mathbb E_{\mathcal D'} \left[ -\frac{1}{|\mathcal D'|} \sum_{x_i',y_i' \in \mathcal D'} -\log f[x_i'](y_i') q(\mathcal E=e_k|x_i',y_i') \right] 
    \right.\right.\\
    &\left.\left. \quad\quad\quad\quad\quad\quad\quad - \frac{1}{|\mathcal D|} \sum_{x_i,y_i \in \mathcal D} -\log f[x_i](y_i) q(\mathcal E=e_k|x_i,y_i) \right| \right] \\
    &\leq \mathbb E_{\mathcal D}\left[ \sup_{f\in \mathcal V, q \in \mathcal Q} \mathbb E_{\mathcal D'} \left| -\frac{1}{|\mathcal D'|} \sum_{x_i',y_i' \in \mathcal D'} -\log f[x_i'](y_i') q(\mathcal E=e_k|x_i',y_i') \right.\right. \\
    &\left.\left.\quad\quad\quad\quad\quad\quad\quad\quad\quad - \frac{1}{|\mathcal D|} \sum_{x_i,y_i \in \mathcal D} -\log f[x_i](y_i) q(\mathcal E=e_k|x_i,y_i) \right| \right] \\
    \label{equ:err_1}
    &\leq \mathbb E_{\mathcal D, \mathcal D'} \left[  \sup_{f\in \mathcal V, q \in \mathcal Q} \frac{1}{|\mathcal D|} \left| \sum_{x_i,y_i \in \mathcal D} \log f[x_i](y_i) q(\mathcal E=e_k|x_i,y_i) \right.\right.\\
    &\left.\left.\quad\quad\quad\quad\quad\quad\quad\quad\quad - \sum_{x_i',y_i' \in \mathcal D'} \log f[x_i'](y_i') q(\mathcal E=e_k|x_i',y_i') \right| \right] \\
    \label{equ:err_2}
    &\left.\left.\quad\quad\quad\quad\quad\quad\quad\quad\quad - \sum_{x_i',y_i' \in \mathcal D'} \sigma_i \log f[x_i'](y_i') q(\mathcal E=e_k|x_i',y_i') \right| \right] \\
    &\leq \mathbb E_{\mathcal D, \sigma} \left[ \sup_{f\in \mathcal V, q \in \mathcal Q} \frac{1}{|\mathcal D|} \left| \sum_{x_i,y_i \in \mathcal D} \sigma_i \log f[x_i](y_i) q(\mathcal E=e_k|x_i,y_i) \right| \right] \\
    &\quad\; + \mathbb E_{\mathcal D', \sigma} \left[ \sup_{f\in \mathcal V, q \in \mathcal Q} \frac{1}{|\mathcal D'|} \left| \sum_{x_i',y_i' \in \mathcal D'} \sigma_i \log f[x_i'](y_i') q(\mathcal E=e_k|x_i',y_i') \right| \right] \\
    \label{equ:err_6}
    &= 2\mathscr R_{|\mathcal D|}(\mathcal G_{\mathcal V}^k),
\end{align}
\end{small}
where $\sigma_i$ are independent Rademacher variables. Equation \ref{equ:err_1} follows from Jensen's inequality and the convexity of $\sup$. Equation \ref{equ:err_2} holds due to the symmetry of $\log f[x_i](y_i) q(\mathcal E=e_k|x_i,y_i) - \log f[x_i'](y_i') q(\mathcal E=e_k|x_i',y_i')$ and the argument that Radamacher variables preserve the expected sum of symmetric random variables with a convex mapping (\cite{banach_probability}, Lemma 6.3).

Substituting Equation \ref{equ:err_6} to Equation \ref{equ:err_7}, we have for any $\delta \in (0,1)$, with a probability over $1 - \delta$, $\forall f \in \mathcal V$, $\forall q \in \mathcal Q$, the following holds:
\begin{align}
\label{equ:err_3}
    &\quad\; \left| q(\mathcal E=e_k)\mathbb E_{q} \left[ -\log f[x](y)|\mathcal E=e_k \right]  - \frac{1}{|\mathcal D|} \sum_{x_i,y_i \in \mathcal D} -\log f[x_i](y_i) q(\mathcal E=e_k|x_i,y_i)   \right|\\
    &\leq 2\mathscr R_{|\mathcal D|}(\mathcal G_{\mathcal V}^k) + B\sqrt{\frac{2\log{\frac{1}{\delta}}}{|\mathcal D|}}.
\end{align}
Let $\Tilde{f_k} = \arg \inf_f \{q(\mathcal E=e_k)\mathbb E_{q} \left[ -\log f[x](y)|\mathcal E=e_k \right]\}$. 

Let $\hat{f_k} = \arg \inf_f \{\frac{1}{|\mathcal D|} \sum_{x_i,y_i \in \mathcal D} -\log f[x_i](y_i)  q(\mathcal E=e_k|x_i,y_i)\}$.

Now we have
\begin{align}
    \label{equ:err_4}
    &\quad\; q(\mathcal E=e_k)\mathbb E_{q} \left[ -\log \Tilde{f_k}[x](y)|\mathcal E=e_k \right]  - \frac{1}{|\mathcal D|} \sum_{x_i,y_i \in \mathcal D} -\log \Tilde{f_k}[x_i](y_i) q(\mathcal E=e_k|x_i,y_i) \\
    &\leq q(\mathcal E=e_k)H_{\mathcal V}(Y|X,\mathcal E=e_k)  - \frac{1}{|\mathcal D|} \sum_{x_i,y_i \in \mathcal D} -\log \hat{f_k}[x_i](y_i) q(\mathcal E=e_k|x_i,y_i) \\
    \label{equ:err_5}
    &\leq q(\mathcal E=e_k)\mathbb E_{q} \left[ -\log \hat{f_k}[x](y)|\mathcal E=e_k \right]  - \frac{1}{|\mathcal D|} \sum_{x_i,y_i \in \mathcal D} -\log \hat{f_k}[x_i](y_i) q(\mathcal E=e_k|x_i,y_i).
\end{align}
Combining Equation \ref{equ:err_3} and Equation \ref{equ:err_4}-\ref{equ:err_5}, the lemma is proved.
\end{proof}

\begin{lemma}
\label{lemma:err_2}
Assume $\forall x \in \mathcal X$,$\forall y \in \mathcal Y$,$\forall f \in \mathcal V$, $\log f[\emptyset](y) \in [-B,B]$ where $B > 0$. The definition of $\mathcal G_{\mathcal V}^k$ and $\mathscr R_{N}(\mathcal G)$ follows from Lemma \ref{lemma:err_1}. Define $\hat f_k = \arg \inf_f \frac{1}{|\mathcal D|}  \sum_{x_i,y_i \in \mathcal D} -\log f[\emptyset](y_i) q(\mathcal E=e_k|x_i,y_i)$.

Then for any $q \in \mathcal Q$,  any $\delta \in (0,1)$, with a probability over $1 - \delta$,  we have
\begin{align}
    &\quad\; \left|q(\mathcal E=e_k)H_{\mathcal V}(Y|\mathcal E=e_k)  - \frac{1}{|\mathcal D|} \sum_{x_i,y_i \in \mathcal D} -\log \hat{f_k}[\emptyset](y_i) q(\mathcal E=e_k|x_i,y_i) \right| \\
    &\leq 2\mathscr R_{|\mathcal D|}(\mathcal G_{\mathcal V}^k) + B\sqrt{\frac{2\log{\frac{1}{\delta}}}{|\mathcal D|}}.
\end{align}
\end{lemma}
\begin{proof}
Similar to  Lemma \ref{lemma:err_1}, we could prove that
\begin{align}
    \label{equ:err_8}
    &\quad\; \left|q(\mathcal E=e_k)H_{\mathcal V}(Y|\mathcal E=e_k)  - \frac{1}{|\mathcal D|} \sum_{x_i,y_i \in \mathcal D} -\log \hat{f_k}[\emptyset](y_i) q(\mathcal E=e_k|x_i,y_i) \right| \\
    &\leq 2\mathscr R_{|\mathcal D|}(\mathcal G_{\mathcal V^\emptyset}^k) + B\sqrt{\frac{2\log{\frac{1}{\delta}}}{|\mathcal D|}},
\end{align}
where $\mathcal G_{\mathcal V^\emptyset}^k = \{g|g(x,y) = \log f[\emptyset](y)q(\mathcal E=e_k|x,y), f\in \mathcal V, q \in \mathcal Q  \}$.

According to the definition for the predictive family $\mathcal V$ (\cite{DBLP:conf/iclr/XuZSSE20}, Definition 1), $\forall f \in \mathcal V$, there exists $f' \in \mathcal V$ such that $\forall x \in \mathcal X$, $f[\emptyset] = f'[x]$. Thus, $\mathcal G_{\mathcal V^\emptyset}^k \subset \mathcal G_{\mathcal V}^k$, and therefore $\mathscr R_{|\mathcal D|}(\mathcal G_{\mathcal V^\emptyset}^k) \leq \mathscr R_{|\mathcal D|}(\mathcal G_{\mathcal V}^k)$. Substituting into Equation \ref{equ:err_8}, the lemma is proved.
\end{proof}

\begin{lemma}[\citep{DBLP:conf/iclr/XuZSSE20}, Theorem 1]
\label{lemma:err_3}
Assume $\forall x \in \mathcal X$,$\forall y \in \mathcal Y$,$\forall f \in \mathcal V$, $\log f[x](y) \in [-B,B]$ where $B > 0$. Define a function class $\mathcal G_{\mathcal V}^* = \{g|g(x,y) = \log f[x](y), f\in \mathcal V\}$. The definition of $\mathscr R_{N}(\mathcal G)$ follows from Lemma \ref{lemma:err_1}. 

Then for any $\delta \in (0,0.5)$, with a probability over $1 - 2\delta$,  we have
\begin{align}
    \left|\mathbb I_{\mathcal V}(X\rightarrow Y)  -  \hat{\mathbb I}_{\mathcal V}(X\rightarrow Y) \right| 
    \leq 4\mathscr R_{|\mathcal D|}(\mathcal G_{\mathcal V}^*) + 2B\sqrt{\frac{2\log{\frac{1}{\delta}}}{|\mathcal D|}}.
\end{align}
\end{lemma}

Finally we are prepared to prove Theorem \ref{theorem:pac}.



\begin{proof}[Proof of Theorem \ref{theorem:pac}]
We first bound the error of empirical estimation with the sum of items in Lemma \ref{lemma:err_1},\ref{lemma:err_2},\ref{lemma:err_3}.
\begin{small}
\begin{align}
    &\quad\; |\mathcal H_\mathcal V^{\mathscr E_K}(X\rightarrow Y) - \hat{H}_\mathcal V^{\mathscr E_K}(X\rightarrow Y;\mathcal D)| \\
    &\leq \left|\sup_{\mathcal E \in \mathscr E_K} \mathbb{I}_{\mathcal{V}}(X\rightarrow Y| {\mathcal{E}}) - \sup_{\mathcal E \in \mathscr E_K} \hat{\mathbb{I}}_{\mathcal{V}}(X\rightarrow Y| {\mathcal{E}};\mathcal D)  \right| 
    + \left| \mathbb{I}_{\mathcal{V}}(X\rightarrow Y) - \hat{\mathbb{I}}_{\mathcal{V}}(X\rightarrow Y;\mathcal D)  \right| \\
    &\leq \sup_{\mathcal E \in \mathscr E_K} \left| \mathbb{I}_{\mathcal{V}}(X\rightarrow Y| {\mathcal{E}}) - \hat{\mathbb{I}}_{\mathcal{V}}(X\rightarrow Y| {\mathcal{E}};\mathcal D) \right| 
    + \left| \mathbb{I}_{\mathcal{V}}(X\rightarrow Y) - \hat{\mathbb{I}}_{\mathcal{V}}(X\rightarrow Y;\mathcal D)  \right| \\
    &= \sup_{q\in \mathcal Q}\left| \sum_{k=1}^K \left[{q}(\mathcal E=e_k)H_{\mathcal V}(Y|\mathcal E=e_k) - {q}(\mathcal E=e_k)H_{\mathcal V}(Y|X, \mathcal E=e_k)\right] \right. \\
    &\left. \quad\quad - \sum_{k=1}^K \left[ q(\mathcal E=e_k)\hat H_{\mathcal V}(Y|\mathcal E=e_k;\mathcal D) - q(\mathcal E=e_k)\hat H_{\mathcal V}(Y|X, \mathcal E=e_k;\mathcal D)\right] \right| \\
    &\quad + \left| \mathbb{I}_{\mathcal{V}}(X\rightarrow Y) - \hat{\mathbb{I}}_{\mathcal{V}}(X\rightarrow Y;\mathcal D)  \right| \\ 
    &\leq \sum_{k=1}^K \sup_{q\in \mathcal Q}\left| {q}(\mathcal E=e_k)H_{\mathcal V}(Y|\mathcal E=e_k) -  q(\mathcal E=e_k)\hat H_{\mathcal V}(Y|\mathcal E=e_k;\mathcal D) \right| \\
    &\quad + \sum_{k=1}^K \sup_{q \in \mathcal Q}\left| {q}(\mathcal E=e_k)H_{\mathcal V}(Y|X, \mathcal E=e_k) -  q(\mathcal E=e_k)\hat H_{\mathcal V}(Y|X, \mathcal E=e_k;\mathcal D) \right| \\
    &\quad + \left| \mathbb{I}_{\mathcal{V}}(X\rightarrow Y) - \hat{\mathbb{I}}_{\mathcal{V}}(X\rightarrow Y;\mathcal D)  \right| \\ 
    &= \sum_{k=1}^K \sup_{q\in \mathcal Q}\left| {q}(\mathcal E=e_k)H_{\mathcal V}(Y|\mathcal E=e_k) - \frac{1}{|\mathcal D|} \sum_{x_i,y_i \in \mathcal D} -\log \hat{f}_k[x_i](y_i) q(\mathcal E=e_k|x_i,y_i) \right| \\
    &\quad + \sum_{k=1}^K \sup_{q \in \mathcal Q}\left| {q}(\mathcal E=e_k)H_{\mathcal V}(Y|X, \mathcal E=e_k) 
    - \frac{1}{|\mathcal D|} \sum_{x_i,y_i \in \mathcal D} -\log \hat{f}_k'[\emptyset](y_i) q(\mathcal E=e_k|x_i,y_i) \right| \\
    &\quad + \left| \mathbb{I}_{\mathcal{V}}(X\rightarrow Y) - \hat{\mathbb{I}}_{\mathcal{V}}(X\rightarrow Y;\mathcal D)  \right|,
\end{align}
\end{small}
where $\hat f_k = \arg \inf_f \frac{1}{|\mathcal D|}  \sum_{x_i,y_i \in \mathcal D} -\log f[x_i](y_i) q(\mathcal E=e_k|x_i,y_i)$,

and $\hat f_k' = \arg \inf_f \frac{1}{|\mathcal D|}  \sum_{x_i,y_i \in \mathcal D} -\log f[\emptyset](y_i) q(\mathcal E=e_k|x_i,y_i)$, for any $q\in \mathcal Q$ and $1\leq k \leq K$. 

For simplicity, let
\begin{small}
\begin{align}
    \mathrm {Err}_k &= \sup_{q \in \mathcal Q}\left| {q}(\mathcal E=e_k)H_{\mathcal V}(Y|X, \mathcal E=e_k) 
    - \frac{1}{|\mathcal D|} \sum_{x_i,y_i \in \mathcal D} -\log \hat{f}_k[x_i](y_i) q(\mathcal E=e_k|x_i,y_i) \right|. \\
    \mathrm {Err}_k' &= \sup_{q \in \mathcal Q}\left| {q}(\mathcal E=e_k)H_{\mathcal V}(Y|X, \mathcal E=e_k) 
    - \frac{1}{|\mathcal D|} \sum_{x_i,y_i \in \mathcal D} -\log \hat{f}_k'[\emptyset](y_i) q(\mathcal E=e_k|x_i,y_i) \right|. \\
    \mathrm {Err}^* &= \left| \mathbb{I}_{\mathcal{V}}(X\rightarrow Y) - \hat{\mathbb{I}}_{\mathcal{V}}(X\rightarrow Y;\mathcal D)  \right|.
\end{align}
\end{small}

Then, by Lemma \ref{lemma:err_1},\ref{lemma:err_2},\ref{lemma:err_3}, 
\begin{small}
\begin{align}
    &\quad\; \mathrm{Pr}\left[|\mathcal H_K^\mathcal V - \hat{\mathcal H}_K^\mathcal V(\mathcal D)| > 4(K+1)\mathscr R_{|\mathcal D|}(\mathcal G_{\mathcal V}) + 2(K+1)B\sqrt{\frac{2\log{\frac{1}{\delta}}}{|\mathcal D|}}\right] \\
    &\leq \mathrm{Pr}\left[\sum_{i=1}^K \mathrm {Err}_k + \sum_{i=1}^K \mathrm {Err}_k' + \mathrm {Err}^* > 4(K+1)\mathscr R_{|\mathcal D|}(\mathcal G_{\mathcal V}) + 2(K+1)B\sqrt{\frac{2\log{\frac{1}{\delta}}}{|\mathcal D|}}\right] \\
    \label{equ:err_9}
    &\leq  \mathrm{Pr}\left[\sum_{i=1}^K \mathrm {Err}_k + \sum_{i=1}^K \mathrm {Err}_k' + \mathrm {Err}^* > \sum_{k=1}^K 4\mathscr R_{|\mathcal D|}(\mathcal G_{\mathcal V}^k) + 4\mathscr R_{|\mathcal D|}(\mathcal G_{\mathcal V}^*) + 2(K+1)B\sqrt{\frac{2\log{\frac{1}{\delta}}}{|\mathcal D|}}\right] \\
    &\left. \quad\quad\quad + \left(\mathrm{Err}^* > 4\mathscr R_{|\mathcal D|}(\mathcal G_{\mathcal V}^*) + 2B\sqrt{\frac{2\log{\frac{1}{\delta}}}{|\mathcal D|}} \right) \right] \\
    &\quad\; + \mathrm{Pr}\left[ \mathrm{Err}^* > 4\mathscr R_{|\mathcal D|}(\mathcal G_{\mathcal V}^*) + 2B\sqrt{\frac{2\log{\frac{1}{\delta}}}{|\mathcal D|}}  \right] \\
    &\leq 2(K+1)\delta.
\end{align}
\end{small}
Equation \ref{equ:err_9} is because of $\mathcal G_{\mathcal V}^k = \mathcal G_{\mathcal V}$, $\mathcal G_{\mathcal V}^* \subset \mathcal G_{\mathcal V}$ and therefore $R_{|\mathcal D|}(\mathcal G_{\mathcal V}^k) \leq R_{|\mathcal D|}(\mathcal G_{\mathcal V})$, $R_{|\mathcal D|}(\mathcal G_{\mathcal V}^*) \leq R_{|\mathcal D|}(\mathcal G_{\mathcal V})$.
Hence,
\begin{small}
\begin{align}
    &\quad \mathrm{Pr}\left[|\mathcal H_\mathcal V^{\mathscr E_K}(X\rightarrow Y) - \hat{H}_\mathcal V^{\mathscr E_K}(X\rightarrow Y;\mathcal D)| \leq 4(K+1)\mathscr R_{|\mathcal D|}(\mathcal G_{\mathcal V}) + 2(K+1)B\sqrt{\frac{2\log{\frac{1}{\delta}}}{|\mathcal D|}}\right] \\
    &\geq 1 - 2(K+1)\delta.
\end{align}
\end{small}
\end{proof}

\section{Proof of Theorem \ref{theorem:IM}}
\label{proof: IM}
\begin{proof}[Proof of Theorem \ref{theorem:IM}]
    The objective function of our IM algorithm is directly derived from the definition of empirical predictive heterogeneity in Definition \ref{def:empirical_predictive_heterogeneity}.
    For the regression task, we assume the predictive family as 
    \begin{small}
\begin{equation}
	\mathcal{V}_1 = \{g: g[x]=\mathcal{N}(f_{\theta}(x), \sigma^2), f\text{ is the regression model and }\theta\text{ is learnable, }\sigma=1.0 (\text{fixed})\},
\end{equation}	
\end{small}
where we only care about the output of the model and the noise scale of the Gaussian distribution is often ignored, for which we simply set $\sigma=1.0$ as a fixed term.
Then for each environment $e\in\text{supp}(\mathcal{E}^*)$, the $\mathbb{I}_{\mathcal{V}}(X\rightarrow Y|\mathcal{E}^*=e)$ becomes
\begin{equation}
	\mathbb{I}_{\mathcal{V}}(X\rightarrow Y|\mathcal{E}^*=e)\propto \min_\theta \mathbb{E}^[\|Y-f_\theta(X)\|^2|\mathcal{E}^*=e] - \text{Var}(Y|\mathcal{E}^*),
\end{equation}
which corresponds with the MSE loss and the proposed regularizer in Equation \ref{equ:regularizer-regression}.
For the classification task, the derivation is similar, and the regularizer becomes the entropy of $Y$ in sub-population $e$ and the loss function becomes the cross-entropy loss.
\end{proof}

\vskip 0.2in
\bibliography{sample}

\end{document}